\documentclass[journal]{IEEEtran}

\usepackage[utf8]{inputenc} 
\usepackage{hyperref}       
\usepackage{url}            
\usepackage{booktabs}       
\usepackage[T1]{fontenc}    
\usepackage{amsfonts}       
\usepackage{nicefrac}       
\usepackage{microtype}      
\usepackage{bm}
\usepackage{amsmath, amssymb}
\usepackage{amsfonts}
\usepackage{amsthm}
\usepackage{xcolor}
\usepackage{stmaryrd}
\usepackage{caption}
\usepackage{subcaption}
\usepackage{graphicx}
\usepackage{bbm}
\usepackage{wrapfig}
\usepackage[normalem]{ulem}
\usepackage{cite}
\usepackage{textcomp}

\usepackage{todonotes}

\newcommand{\minimize}[1]{\underset{#1}{\text{minimize }}}
\newcommand{\R}{\mathbb{R}}

\def\x{{\bm{x}}}
\def\r{{\bm{r}}}
\def\v{{\bm{v}}}
\def\th{{\bm{\theta}}}
\def\L{{\mathcal{L}}}
\def\sota{{state-of-the-art }}

\begin{document}

\title{Optimism in the Face of Adversity: \\  Understanding and Improving Deep Learning through Adversarial Robustness}
    
\author{Guillermo~Ortiz-Jim\'enez,
        Apostolos~Modas,
        Seyed-Mohsen~Moosavi-Dezfooli,
        and~Pascal~Frossard
\thanks{This work was supported in part by the {CHIST-ERA} program through the project CORSMAL, under Swiss NSF grant 20CH21{\_}180444.}
\thanks{G. Ortiz-Jim\'enez, A. Modas and P. Frossard are with Ecole Polytechnique F\'ed\'erale de Lausanne (EPFL), Switzerland. Email: \{guillermo.ortizjimenez,apostolos.modas,pascal.frossard\}@epfl.ch}
\thanks{S.M. Moosavi-Dezfooli is with the ETH Z\"urich, Switzerland. Email: seyed.moosavi@inf.ethz.ch}
}

\markboth{Preprint}%
{Optimism in the Face of Adversity in Deep Learning}
%



\maketitle

\begin{abstract}
Driven by massive amounts of data and important advances in computational resources, new deep learning systems have achieved outstanding results in a large spectrum of applications. Nevertheless, our current theoretical understanding on the mathematical foundations of deep learning lags far behind its empirical success. However, the field of adversarial robustness has recently become one of the main sources of explanations of our deep models. In this article, we provide an in-depth review of the field, and give a self-contained introduction to its main notions. But, in contrast to the mainstream pessimistic perspective of adversarial robustness, we focus on the main positive aspects that it entails. We highlight the intuitive connection between adversarial examples and the geometry of deep neural networks, and eventually explore how the geometric study of adversarial examples can serve as a powerful tool to understand deep learning. Furthermore, we demonstrate the broad applicability of adversarial robustness, providing an overview of the main emerging applications of adversarial robustness beyond security. The goal of this article is to provide readers with a set of new perspectives to understand deep learning, and to supply them with intuitive tools and insights on how to use adversarial robustness to improve it.
\end{abstract}


%
\IEEEpeerreviewmaketitle

\section{Introduction}

\IEEEPARstart{A}{lready} back in the 1940s, it was suggested that neurally inspired algorithms could serve as powerful general purpose machine learning models~\cite{mcculloch_pitts_1943}. Nevertheless, it was not until very recently, that artificial neural networks started to become the \sota in most machine learning benchmarks with the emergence of the field of deep learning~\cite{lecun_bengio_hinton_2015}. Propelled by a new abundance of data and important advances in hardware design, new deep learning systems have achieved outstanding results in a wide range of applications, even performing on par with human baselines in certain benchmarks. For example, the accuracy on image recognition benchmarks has steadily grown in recent years~\cite{EvaluatingImageNet}, and computer vision applications based on deep learning now permeate the consumer market. Also, artificial neural networks can now compete with the best humans at certain complex games~\cite{alphaGO_2016}, while the quality of language generation and understanding is swiftly reaching exciting levels~\cite{gpt3}.

Deep learning is based on a simple principle. A deep neural network architecture is defined as the composition of several simple functions, or layers, and trained end-to-end using standard optimization algorithms. The philosophy behind this principle resides in the idea that a proper choice of architecture, alongside vasts amounts of training data, can lead a neural network to learn good representations of the training distribution.

Nevertheless, keeping up with a rapid and steady improvement of the performance of deep learning systems has come at a cost. Our current theoretical understanding on the mathematical foundations of deep learning lags far behind its empirical success. Understanding deep learning is, indeed,  hard: the functional relationship between different layers, or explaining its dynamics in a high-dimensional and non-convex setting, lie outside of the scope of our current mathematical knowledge. As a result, even if deep learning often works well in practice, we still cannot completely answer why, or how, it does so.

A new line of research has, therefore, emerged trying to explain and understand the fundamental mechanisms behind deep learning. The role of research in this area is to explain the large empirical evidence and support it with new theories. Nevertheless, for theory to follow evidence, multiple experiments were carried out to test the limits of these systems. In this sense, it was expected that, because neural networks excel at many tasks, they would be guided by similar rules and abstractions as the humans with which they compete. In fact, the common folklore in deep learning was that as a data sample goes through a neural network, the layers in the architecture extract most of its information, and compress it in a hierarchical and abstract representation.

\begin{figure}[t]
    \centering
    \includegraphics[width=\columnwidth]{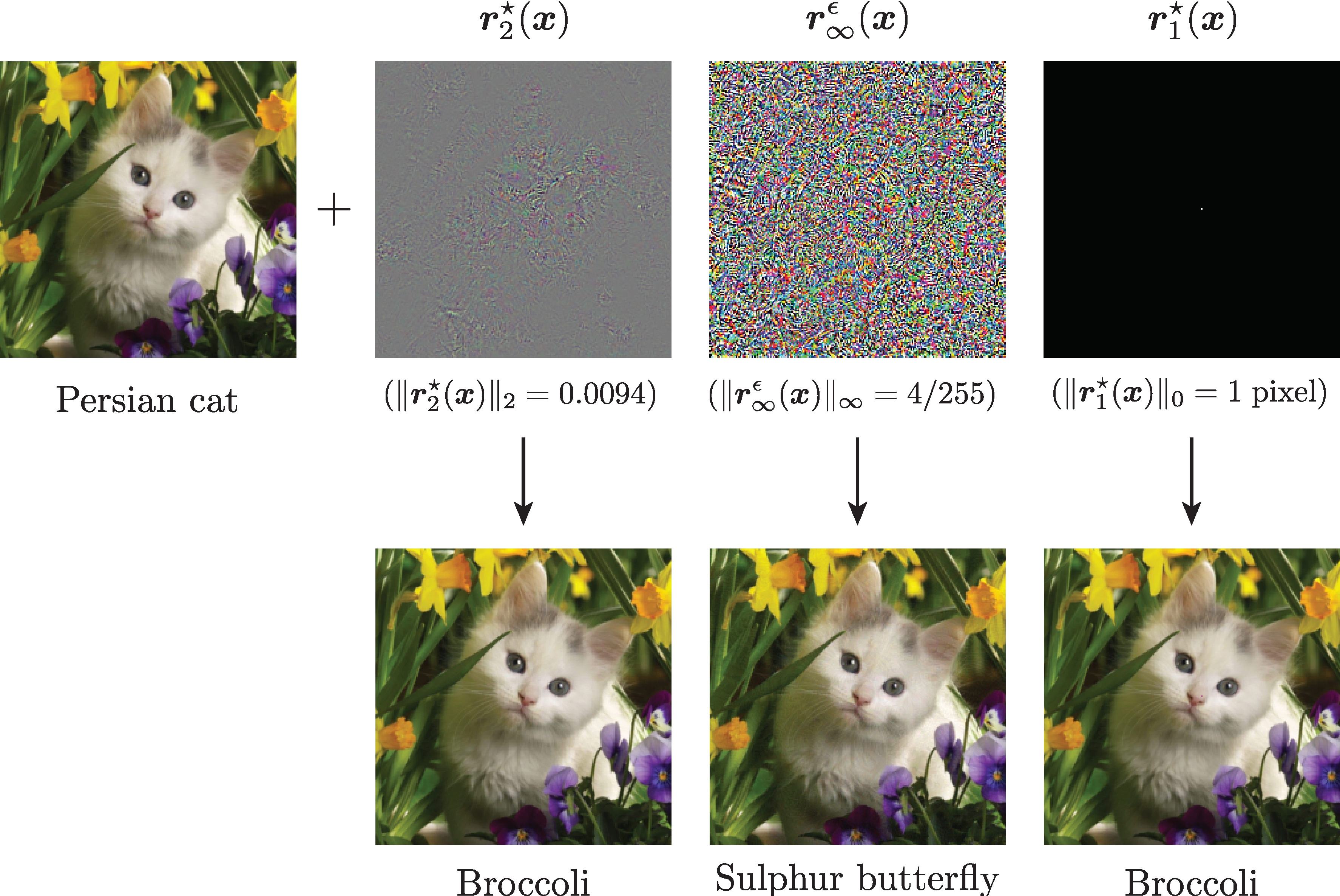}
    \caption{Illustration of different additive adversarial perturbations ($\ell_1, \ell_2$ and $\ell_\infty$) (top) and the corresponding adversarial examples (bottom) that fool a \sota deep neural network. The norm of each perturbation is indicated below the corresponding image, except for the $\ell_1$ perturbation (sparse) where the number of perturbed pixels is provided. The perturbations are computed on a VGG-16~\cite{VGG} trained on ImageNet~\cite{dengImageNet}, and the resulting misclassified labels are shown below each adversarial example. In all cases the adversarial example is hardly distinguishable by a human observer. The original image is taken from the web.}
    \label{fig:adv_examples}
\end{figure}

This belief, however, has been heavily challenged with the discovery that \emph{adversarial examples}~\cite{AdversarialClassification, lowd_adversarial_learning} are also ubiquitous in deep learning~\cite{szegedyIntriguingPropertiesNeural2014}. Indeed, for virtually any sample that is correctly classified by a neural network, you can always add some (imperceptible) perturbation, which, although it does not affect human recognition, it utterly changes the prediction of the deep classifier (see Figure~\ref{fig:adv_examples}).This has important ramifications in many applications of deep learning, as it raises serious concerns about the safety, trustworthiness, and fairness of deep neural networks. This intriguing phenomenon, alongside the shortcomings of deep learning theory, has made the study of adversarial examples one of the main issues of the deep learning research agenda hitherto.

The research community has been mostly preoccupied so far with the security risks of adversarial examples.
Indeed, the fact that neural networks are so sensitive to adversarial perturbations makes them vulnerable to attacks, when they are deployed in hostile environments. Making neural networks robust against these attacks is, therefore, a security issue. In this sense, what started simply as a method to craft these adversarial perturbations, suddenly became a fierce battle between proponents of new methods to fool these systems (attacks) and those proponents of security patches (defenses)~\cite{biggio_wild_2018,AkhtarSurvey,yuan_adversarial_2019,miller_adversarial_2020}. The philosophy behind this approach is simple: to make these models safe from adversarial examples, we need to defend them against the strongest and most sophisticated attacks. In this attack-defense cycle, it was found that virtually all applications of deep networks suffer from adversarial vulnerabilities, while making these networks robust is a hard challenge.

Towards building more robust models, the field of \emph{adversarial robustness} has had to push the limits of deep learning. Mainly because designing novel attacks and stronger defenses is a challenging task. A deep classifier assigns a label to any input sample, and, as such, it defines a partition of the input space in regions with different labels. The problem of finding adversarial examples is therefore reduced to the problem of finding close data points that belong to different classification regions (see Figure~\ref{fig:diagram_adv}). Researchers in adversarial robustness have had, hence, to derive new tools to explore the geometry of the decisions of neural networks in the input space~\cite{fawziRobustnessDeepNetworks2017}. 

Fortunately enough, although designed from a security perspective, these geometric tools rapidly turned out to be useful to analyze other properties of deep neural networks. Indeed, adversarial examples explicitly manipulate the decisions of a neural network, and as such, they reveal the functional properties of deep learning systems. As a consequence, the adversarial robustness field has provided many insights to our understanding of deep learning.

Besides, adversarially robust models do not suffer from the same vulnerabilities as standard deep neural networks, and are, therefore, more reliable. Recent studies suggest that robust neural networks are easier to understand from a human perspective~\cite{rossImprovingAdversarialRobustness,tsiprasRobustnessMayBe2018}, in the sense that the internal representations of robust models are seemingly closer to our human intuitions. If this property were to be confirmed, this would alleviate some concerns regarding the trust in deep learning systems. In fact, a great deal of research currently relies on this hypothesis and tries to exploit adversarially robust models in applications that require greater levels of invariance and stronger representations. This is proving to be a fertile ground with promising results.

In this article, we provide an in-depth review on the benefits of adversarial robustness in deep learning, and give a self-contained introduction to the main notions of the field. In particular, we start with an overview of the fundamental concepts in adversarial robustness, highlighting their intuitive connection with the geometry of deep neural networks. By doing this, we explore how the geometric study of adversarial examples can serve as a powerful tool to understand deep neural networks. In this sense, we describe the connections between adversarial examples and generalization in deep neural networks, and the important links between robustness and the dynamics of learning. Finally, we give an overview of the main emerging applications of adversarial robustness beyond security. In this article, we do not intend to give a comprehensive survey of the \sota in adversarial attacks and defenses, nor give a thorough list of all the applications of adversarial robustness in deep learning. Instead, we focus on providing the readers with a set of new perspectives to understand deep learning, and on supplying them with intuitive tools and insights on how to use adversarial robustness to improve it.

The rest of this article is structured as follows: In Section~\ref{sec:geometric_view}, we review the fundamental ideas and concepts in adversarial robustness through a geometric point of view. Following this, in Section~\ref{sec:understanding_dl}, we use those concepts to analyze the intersection of adversarial robustness and deep learning theory and its main contributions to our understanding of deep neural networks. Afterwards, in Section~\ref{sec:applications_robustness}, we show how adversarial robustness can be used to improve the performance of multiple deep learning applications, and provide a few examples of emerging deep learning fields that are currently exploiting adversarial robustness. Finally, in Section~\ref{sec:future}, we highlight the main open questions and future challenges of the field of adversarial robustness. 

\section{A geometric view of adversarial robustness}
\label{sec:geometric_view}
In this section, we review the fundamental ideas and concepts in adversarial robustness through a geometric point of view. This perspective highlights the connection between adversarial examples and the local geometric properties of neural networks, and it provides an intuitive framework to study robustness properties of deep classifiers. We start by reviewing the main relevant concepts in deep learning throughout this article and highlight some intriguing aspects of neural networks (Section~\ref{sec:dl_defs}). We then introduce the definition of adversarial examples and robustness of a classifier (Section~\ref{sec:def_adv}). Afterwards, we illustrate how the study of adversarial examples can reveal important geometric features of the decision boundary of neural networks (Section~\ref{sec:geometric_insights}). We continue showing how these geometric insights can also help to evaluate and improve the robustness properties of deep classifiers (Section~\ref{sec:geometry_applications}). Finally, we end this section with a discussion on what the existence of adversarial examples means for deep learning (Section~\ref{sec:discussion_adv_examples}). 

\subsection{Classification framework}\label{sec:dl_defs}

We first introduce the classification framework that will be considered throughout this paper. This framework has been the most extensively studied one and has served as a benchmark for different adversarial defense/attack algorithms. We use it to introduce the main robustness definitions below. These definitions can however be extended to other settings and machine learning tasks such as regression~\cite{nguyen2018adversarial,gottschling2020troublesome,harford2020adversarial,Antun2019OnInstabilitiesOfDL}, generative models~\cite{KosAdvExForGANs}, or semantic segmentation~\cite{Xie_Segmentation,Arnab_2018_CVPR}. 

\begin{figure}[t]
    \centering
    \includegraphics[width=\columnwidth]{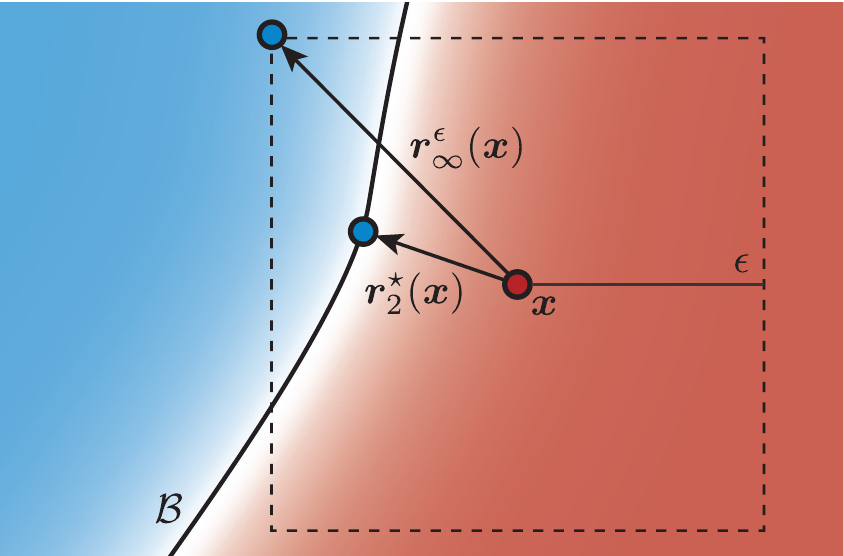}
    \caption{Diagram of the local geometry of a neural network $f_\th$ in the vicinity of a data sample $\x$. The decision boundary $\mathcal{B}$ separates between classification regions with different labels. The colors represent regions with different assigned labels, and the shades the confidence in the prediction. The adversarial perturbations $\r_2^\star(\x)$ and $\r_\infty^\epsilon(\x)$ are two examples of vectors that can move $\x$ to the other side of the boundary.}
    \label{fig:diagram_adv}
\end{figure}

A deep neural network classifier can be described as a function $f_\th:\mathcal{X}\rightarrow\mathcal{Y}$, parameterized by a set of variables, or weights, $\th\in\mathcal{W}$, which maps any input vector $\x\in\mathcal{X}$ to a label $y\in\mathcal{Y}$. We will refer to $\mathcal{W},\mathcal{X}$ and $\mathcal{Y}$ as the weight space, input space, and output space of a neural network, respectively; and without loss of generality we will assume $\mathcal{W}=\R^M$, $\mathcal{X}=\R^D$, and $\mathcal{Y}=\{1,\dots,C\}$.  The role of a classifier is to partition the whole input space into a set of regions with a classification label. The decision boundary $\mathcal{B}$ of the classifier can then be defined as the set of points that lies at the intersection of two classification regions with different labels (see Figure~\ref{fig:diagram_adv}).

Modern feed-forward neural networks~\cite{lecun_bengio_hinton_2015} are formed by the composition of multiple layers where the output $\bm{z}_t\in\R^{D_t}$ of layer $t$ depends only on the output of the previous layers, i.e.,
\begin{equation}
    \bm{z}_t=h_t\left(\bm{z}_{t-1},\bm{z}_{t-2}, \dots, \bm{z}_0;\th_t\right)\quad\text{and}\quad \bm{z}_0=\x.
\end{equation}
Here $h_t:\R^{D_{t-1}}\times\dots\times \R^{D_0}\times\R^{M_t}\rightarrow \R^{D_t}$ can be any general differentiable, or subdifferentiable, mapping parameterized by some weights $\th_t\in\R^{M_t}$. It normally consists of the combination of linear operators (e.g., convolutions) and point-wise non-linearities (e.g., rectified linear units (ReLU)). In classification settings, the entries of the output of the last layer $\bm{z}_L\in\R^{D_L}$ with $D_L=C$ are generally referred to as \emph{logits}. To make the output more interpretable in the classification setting, logits are normally mapped to a set of probabilities $p_\th(\x)\in[0,1]^C$ using a softmax operator, i.e., 
\begin{equation}
    [p_\th(\x)]_k=\cfrac{\exp\left([\bm{z}_L]_k\right)}{\sum_{c=1}^C\exp\left([\bm{z}_L]_c\right)}.
\end{equation}
The predicted class of a neural network classifier is, hence, the index of the highest estimated probability
\begin{equation}
    f_\th(\x)=\underset{k\in\{1,\dots,C\}}{\operatorname{arg\;max}}\; [p_\th(\x)]_k.
\end{equation}

In the supervised setting, given a data distribution $\mathcal{D}$ over pairs $(\x,y)$, the goal of a learning algorithm is to find a classifier $f_\th$ that maps any input $\x$ to a label $y$, such that its \emph{expected risk} on $\mathcal{D}$ is minimized, i.e., 
\begin{equation}
    \min_{\th}\;\mathbb{E}_{(\x,y)\sim\mathcal{D}}\left[\mathcal{L}\left(\x, y; \th\right)\right], \label{eq:accuracy}
\end{equation}
where $\mathcal{L}(\x,y;\th)$ defines a suitable loss function, e.g., cross-entropy, between the logits of a network and the true class label $y$~\cite{Shalev_Shwartz}. In practice, however, we do not have access to the full data distribution $\mathcal{D}$, but, instead, we only know a set of training samples $\{(\x^{(i)},y^{(i)})\}_{i=1}^N\sim\mathcal{D}^N$. As a result, $f_\th$ cannot be obtained by minimizing \eqref{eq:accuracy}, and it is usually obtained as the solution to the \emph{empirical risk} minimization problem
\begin{equation}
    \min_{\th}\;\cfrac{1}{N}\sum_{i=1}^N\;\mathcal{L}\left(\x^{(i)}, y^{(i)}; \th\right). \label{eq:erm}
\end{equation}
In machine learning, the difference between the expected risk in \eqref{eq:accuracy} and the empirical risk in \eqref{eq:erm} attained by a classifier $f_\th$ is known as the \emph{generalization gap} of a classifier~\cite{Shalev_Shwartz}. We say that a neural network achieves good generalization performance when its expected risk is low, and hence its generalization gap is rather small.

As a consequence of their intricate structure and the high-dimensionality of their representations, neural networks can in principle realize arbitrarily complex functions~\cite{cybenko_1989}. Nevertheless, this complexity also makes their behaviour hard to comprehend. In fact, for most neural networks used in practice, a closed-form analysis of their properties, such as the analytical characterization of their classification regions, is not possible with our current mathematical tools. For this reason, understanding the high-dimensional geometry of the decision boundary of these classifiers is a  challenging task. 

\begin{figure}[t]
    \centering
    \includegraphics[width=\columnwidth]{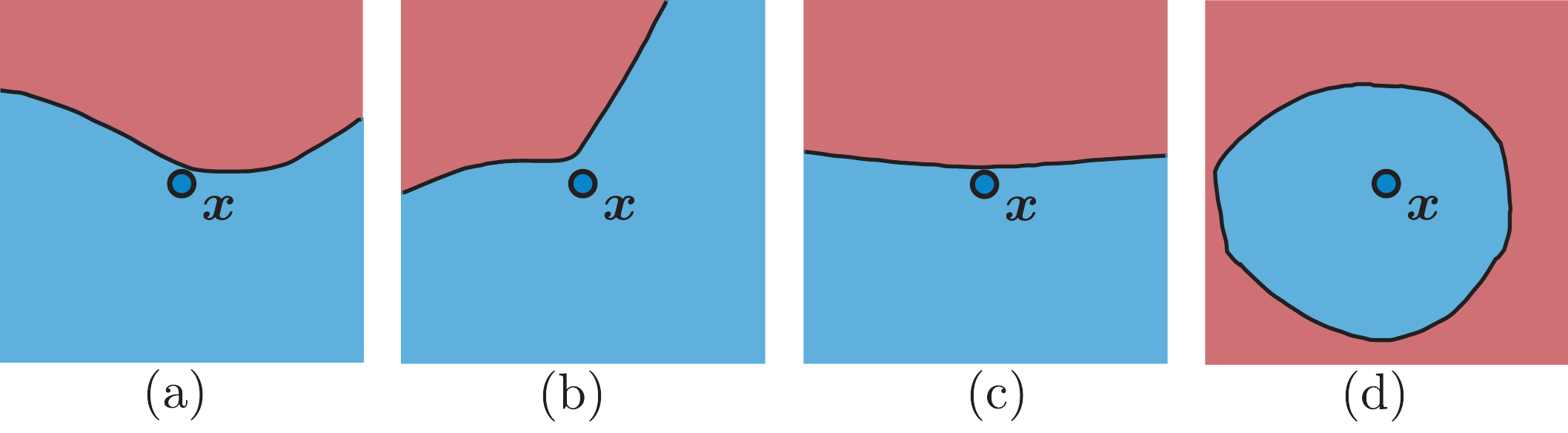}
    \caption{Examples of some cross-sections of the decision boundary of a deep neural network in the vicinity of a data sample $\x$. Depending on the directions used to span the visualization plane we can observe very different geometrical behaviours in the boundary of this classifier. (a) and (b) show cross-sections of a highly curved boundary in the vicinity of $\x$. On the other hand, (c) shows a a flat boundary in the vicinity of $\x$. (d) Shows a cross-section with two random directions, and as a result the boundary is very far from the data point. Both axes are scaled equally in all images. Figure taken from~\cite{fawziRobustnessDeepNetworks2017}; with permission from the authors.}
    \label{fig:cross_sections}
\end{figure}

A useful way to study the properties of the decision boundary of a neural network is to visualize its cross-section with some two-dimensional plane. That is, given two orthonormal vectors $\bm{u}_1,\bm{u}_2\in\mathbb{S}^{D-1}$, such that $\bm{u}_1\perp\bm{u}_2$, and a data sample $\x$ we can visualize the decision boundary of a neural network $f_\th$ in the vicinity of $\x$ by plotting
\begin{equation}
    V(\alpha_1,\alpha_2) = f_\th(\x+\alpha_1\bm{u}_1+\alpha_2\bm{u}_2),
\end{equation}
for some specific values of $\alpha_1$ and $\alpha_2$. For most deep networks, it is common that these cross-sections exhibit very different shapes depending on the choice of $\bm{u}_1$ and $\bm{u}_2$ (see Figure~\ref{fig:cross_sections}). These can range from highly curved to mostly flat boundaries.

Besides, a notable feature of most neural networks trained on high-dimensional datasets is that, in most random cross-sections, the decision boundary appears relatively far from any typical data sample, as shown in Figure~\ref{fig:cross_sections}(d). This empirical observation can be made more rigorous if one studies the robustness of a neural network to \emph{additive random noise}. That is, the probability that a given data sample $\x$, perturbed by a random vector $\v\sim\mathcal{N}(\bm{0},\sigma^2\bm{I})$, is classified different than $\x$, i.e.,
\begin{equation}
    \mathbb{P}_{(\bm{x},y)\sim\mathcal{D}}(f_\th(\x)\neq f_\th(\x+\v)).
\end{equation}
Indeed, for most neural networks used in practice one needs to add noise with a very large variance $\sigma^2$ to fool a classifier~\cite{szegedyIntriguingPropertiesNeural2014}. This is, of course, a desired property of a robust classifier. In fact, this property is also shared by linear classifiers and some kernel methods, for which it can theoretically be shown that such robustness has a positive dependence on the input dimensionality~\cite{fawziRobustnessClassifiersAdversarial2016, Fawzi_Analysis_MachineLearning} (see Sec.~\ref{subsec:lower_bound} for more details). This similarity suggests that, despite their complex structure, neural networks create simple decision boundaries in the vicinity of data samples.

\subsection{Adversarial robustness}\label{sec:def_adv}

Intriguingly, the apparent robustness to random noise described above contrasts with the vulnerability of neural networks to adversarial perturbations~\cite{szegedyIntriguingPropertiesNeural2014}. These are worst-case small modifications of an input $\x$ which are specifically crafted to fool a classifier $f_\th$; random perturbations, on the other hand, are statistically independent from $\x$ or $f_\th$ and are randomly drawn from some noise distribution. Surprisingly, for virtually any $\x$ and $f_\th$ we can always find some adversarial perturbations, which suggests that there always exist some directions for which the decision boundary of a neural network is very close to a given data sample. Hence, adding a very small perturbation in such a direction changes the output of the classifier. 

More formally, we define an \emph{adversarial perturbation} $\r(\x)\in\R^D$ as the solution to the following optimization problem
\begin{equation}
    \begin{split}
        \min_{\r\in\R^D}&\quad Q(\r)\\
        \text{s.t.}&\quad f_\th(\x+\r)\neq f_\th(\x)\\
        &\quad \r\in\mathcal{C}.
    \end{split}
\label{eq:adv_example}
\end{equation}
Here, $Q(\r)$ represents a general objective function, and $\mathcal{C}$ denotes a general set of constraints that characterises the perturbations. We generally refer to the perturbed samples $\x+\r(\x)$ as \emph{adversarial examples} (see Figure~\ref{fig:adv_examples}).

Different types of adversarial perturbations differ in the way that $Q(\r)$ and $\mathcal{C}$ are instantiated\footnote{Note that in most computer vision applications the constraint set $\mathcal{C}$ is generally augmented to also reflect some image range constraints, i.e., $\x+\r\in[0,1]^D$. This is introduced mostly as a security requirement that prevents adversarial examples from being trivially identified based only on the range of their pixel intensities. As security is not the main concern of this review article, and we are also interested in other domains, we will omit the range constraints in most explanations. We advise the readers to check the original articles when interested in the details of a particular implementation.}. For example:
\begin{itemize}
    \item \emph{minimal $\ell_p$ adversarial perturbations}  $\r_p^\star(\x)$ are defined by setting
    \begin{equation}
        Q(\r)=\|\r\|_p=\left(\sum_{k=1}^D ([\r]_k)^p\right)^{1/p},
    \end{equation}
    and $\mathcal{C}=\varnothing$. They represent the notion of the smallest additive perturbation (in an $\ell_p$ sense) required to cross the decision boundary of a classifier~\cite{moosavi-dezfooliDeepFoolSimpleAccurate2016,Carlini_Wagner}.
    \item \emph{$\epsilon$-constrained adversarial perturbations} $\r^\epsilon_p(\x)$ are defined by setting 
    \begin{equation}
        Q(\r)=-\L(\x+\r,y;\bm{\theta}),
    \end{equation}
   and
   \begin{equation}
       \mathcal{C}=\{\r\in\R^D: \|\r\|_p\leq\epsilon\}.\label{eq:epsilon_constraint}
   \end{equation}
   They represent the worst-case perturbation maximizing the loss in a given radius $\epsilon$ around a data sample $\x$. The radius $\epsilon$ is chosen such that the resulting perturbation is small, or even imperceptible~\cite{goodfellowExplainingHarnessingAdversarial2015,madryDeepLearningModels2018}.
    \item Adversarial examples have also been defined using other notions of distance\footnote{Originally, some of these perturbations were not introduced in the form of \eqref{eq:adv_example} as they are not directly defined as additive perturbations. Instead, they define an abstract mapping $\mathcal{T}_{\bm{\phi}}:\R^D\rightarrow\R^D$ parameterized by a set of weights $\bm{\phi}$ that transforms $\x$ into $\mathcal{T}_{\bm{\phi}}(\x)$. Their optimization is, therefore, defined in the parameters $\bm{\phi}$. Nevertheless, non-additive perturbations can always be cast to the form of \eqref{eq:adv_example} by substituting $\r(\x)=\mathcal{T}_{\bm{\phi}}(\bm{x}) - \x$.}, such as geodesics in a data manifold~\cite{fawziManitestAreClassifiers2015, kanbakGeometricRobustnessDeep2018, EngstromExploring}, perceptibility metrics~\cite{Laidlaw_Functional, Sharif_2018_CVPR_Workshops, laidlaw2020perceptual, jordan2019quantifying}, Wasserstein distances~\cite{WongWasserstein, WuWY20} or discrete settings~\cite{zhang_adversarial_2020, yang_greedy_2020}.
\end{itemize}

The $\ell_p$ perturbations are certainly the most commonly studied adversarial perturbations in the literature, and we will mainly focus on those in the remainder of this paper, for the sake of clarity. In Section~\ref{sec:common_corruptions}, however, we will see how exploiting adversarial perturbations beyond the $\ell_p$ metrics can be helpful to tackle a more general notion of robustness.

In the adversarial robustness literature, the algorithms that try to solve \eqref{eq:adv_example} are typically referred to as \emph{adversarial attacks}. We refer the interested reader to~\cite{biggio_wild_2018,AkhtarSurvey,yuan_adversarial_2019,miller_adversarial_2020} for a comprehensive review of the \sota in adversarial attacks. In general, solving \eqref{eq:adv_example} exactly is \emph{a priori} intractable because of the non-convex nature of the classifier $f_\th$ and the high-dimensionality of the input space~\cite{Reluplex}. Hence, most attacks only obtain an approximate solution to this problem. Surprisingly though, such approximate solutions are usually easy to find using first-order optimization methods~\cite{goodfellowExplainingHarnessingAdversarial2015,moosavi-dezfooliDeepFoolSimpleAccurate2016,Carlini_Wagner,madryDeepLearningModels2018}. And, the same way that it is normally sufficient to use stochastic gradient descent to solve \eqref{eq:erm} in the weight space, the optimization of \eqref{eq:adv_example} in the input space using some variant of gradient descent is also effective. As we will see in Section~\ref{sec:geometric_insights}, this is a consequence of the geometry of deep classifiers.

The fact that adversarial examples are so easy to compute, alongside their existence in the first place, exposes a crucial weakness of current state-of-the-art deep classifiers. Towards addressing this issue, it is important to define some target metric that quantifies the susceptibility of a given neural network to adversarial perturbations. Depending on the application, or task, we can define the \emph{adversarial robustness} $\rho(f_\th)$ of a classifier in different ways. One typical formulation is based on the generalization capacity of a classifier in an adversarial setting, and it defines adversarial robustness as the worst-case accuracy of a neural network subject to some adversarial perturbation, i.e.,
\begin{equation}
    \rho_p^\epsilon(f_\th)=\mathbb{P}_{(\x,y)\sim\mathcal{D}}\left(f_\th(\x+\r^\epsilon_p(\x))=y\right). \label{eq:adv_accuracy}
\end{equation}
This quantity is relevant from a security perspective as it highlights the vulnerability of deep neural networks to certain adversarial attacks. In particular, the value of $\epsilon$ -- more generally, the size of the constraint set $\mathcal{C}$ in \eqref{eq:adv_example} -- reflects the strength of the attacker, and, in combination with the choice of metric, e.g., $\ell_p$ norm, determines the \emph{threat model} of an adversary. For the majority of threat models, most standard neural networks have very low worst-case accuracy, in the sense of \eqref{eq:adv_accuracy}, despite their high standard generalization performance as measured by \eqref{eq:accuracy}~\cite{szegedyIntriguingPropertiesNeural2014, moosavi-dezfooliDeepFoolSimpleAccurate2016}. 

Alternatively, one can also measure robustness of a neural network as the average distance of any data sample to the decision boundary of a network, i.e.,
\begin{equation}
    \rho_p^\star(f_\th)=\mathbb{E}_{(\x,y)\sim\mathcal{D}}\left[\|\r_p^\star(\x)\|_p\right]. \label{eq:distance_boundary}
\end{equation}
In this geometric formulation, robustness becomes purely a property of the classifier, as it is independent of the strength of an adversary measured by $\epsilon$ or $\mathcal{C}$. Under this metric, making a classifier more robust means that its boundary is pushed further away from the data samples.

In fact, measuring the ``true'' robustness of a classifier in terms of \eqref{eq:adv_accuracy} is  challenging. The current adversarial attacks are not optimal in computing the perturbation $\r(\x)$ and, consequently, so is the evaluation of \eqref{eq:adv_accuracy} that characterises the training of robust classifiers~\cite{carlini2019OnEvaluating}. Still, one can \emph{verify} the robustness of a classifier by checking if a safe distance exists between its decision boundaries and all data samples. That is, a classifier is certifiably $\epsilon$-robust (in an $\ell_p$-sense) if it always outputs a constant label in an $\ell_p$ ball of radius $\epsilon$ around any typical data sample. One can easily understand though that obtaining such guarantee in a high-dimensional setting requires a considerable computational burden. Indeed, the verification methods proposed in the literature are still computationally expensive, and, for tractability, they are usually constrained to specific type of classifiers~\cite{Reluplex, gowal2019effectiveness, ruan_verification, huang_verification_survey}. However, the field  is currently experiencing significant progress in this regard~\cite{Gowal_2019_ICCV,verification_randomized_smoothing,certification_semidefinite_deepmind}.

Because standard training techniques yield classifiers with good accuracy but low adversarial robustness, making a classifier robust to adversarial perturbations is a complementary objective to generalization performance (see Sec.~\ref{subsec:adversarial_robustness_and_generalization} for a more detailed discussion on the topic). In fact, depending on the application, one should also take into account robustness as an objective during training. For instance, this can be done by replacing \eqref{eq:accuracy} with
\begin{equation}
    \min_{\th}\;\mathbb{E}_{(\x,y)\sim\mathcal{D}}\left[\max_{\r\in\mathcal{C}}\mathcal{L}\left(\x+\r, y; \th\right)\right]. \label{eq:adv_training}
\end{equation}
However, solving \eqref{eq:adv_training} adds yet another level of complexity to the training, as we are replacing a non-convex minimization problem with a non-convex and non-concave minimax optimization problem.

Alternatively, there exist multiple constructive methods that try to improve robustness by maximizing \eqref{eq:adv_accuracy} or \eqref{eq:distance_boundary}. These algorithms are generally known as \emph{adversarial defenses}. Nevertheless, in stark contrast with the success of adversarial attacks, most adversarial defenses are not successful in achieving good levels of robustness~\cite{Carlini_Bypassing,athalyeObfuscatedGradientsGive2018}. In fact, adversarial training~\cite{goodfellowExplainingHarnessingAdversarial2015, moosavi-dezfooliDeepFoolSimpleAccurate2016, madryDeepLearningModels2018}, a method that augments/replaces the training data with adversarial examples crafted during training, is up to this date one of the most effective adversarial defense methods. Nevertheless, its greater computational cost -- sometimes up to fifty times higher than standard training -- makes it impractical for many large-scale applications. For this reason, a great body of literature in adversarial robustness deals with finding alternatives to adversarial training which are at least as robust, but much more computationally efficient. Moreover, it has empirically been shown that performing adversarial training to improve the robustness to a certain type of perturbations, e.g., $\r_\infty^\epsilon(\x)$, may not necessarily improve robustness to other types of attacks, e.g., $\r_1^\star(\x)$~\cite{ModasSparseFool}, making the system vulnerable to other threat models~\cite{Maini_UnionLp,Tramer_MultiplePerts}.

\subsection{Geometric insights of robustness} \label{sec:geometric_insights}

The lack of robustness of deep neural network raises serious concerns in terms of their security and safety in real-life applications like autonomous driving~\cite{modas_toward_2020} or medical imaging~\cite{Finlayson1287}. However, besides exposing these risks, it also reveals important geometric properties of the decision boundaries of deep networks and sheds light onto their structure.

\begin{figure}[t]
\centering
\begin{subfigure}[b]{0.8\columnwidth}
    \includegraphics[width=\columnwidth]{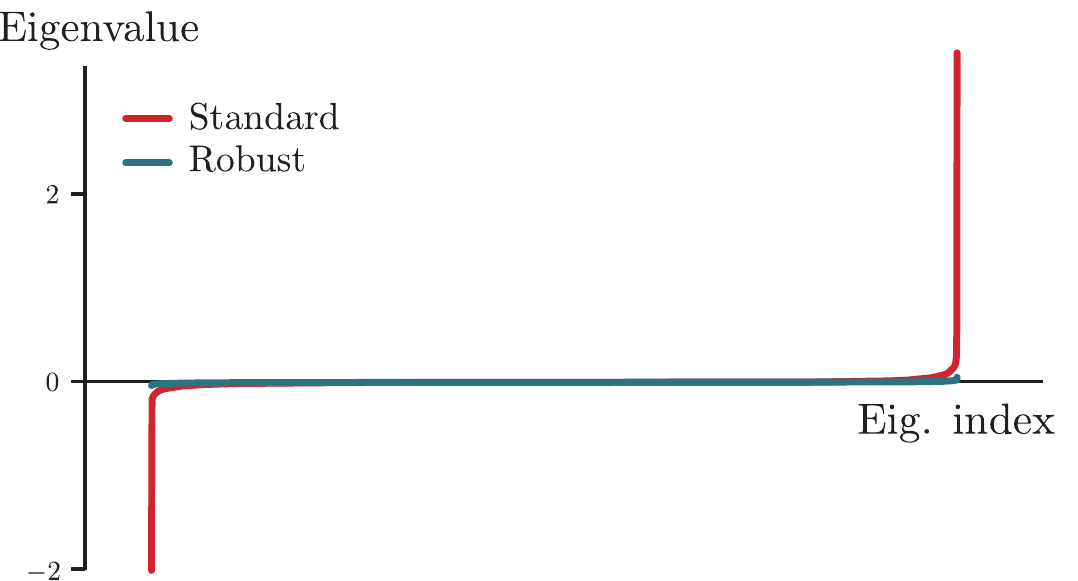}
    \caption{Eigenvalues of $\nabla_\x^2\ell(\x)$.}
    \label{fig:eig_profile}
    \vspace{1.5em}
\end{subfigure}
\begin{subfigure}[b]{0.8\columnwidth}
    \centering
    \includegraphics[width=\columnwidth]{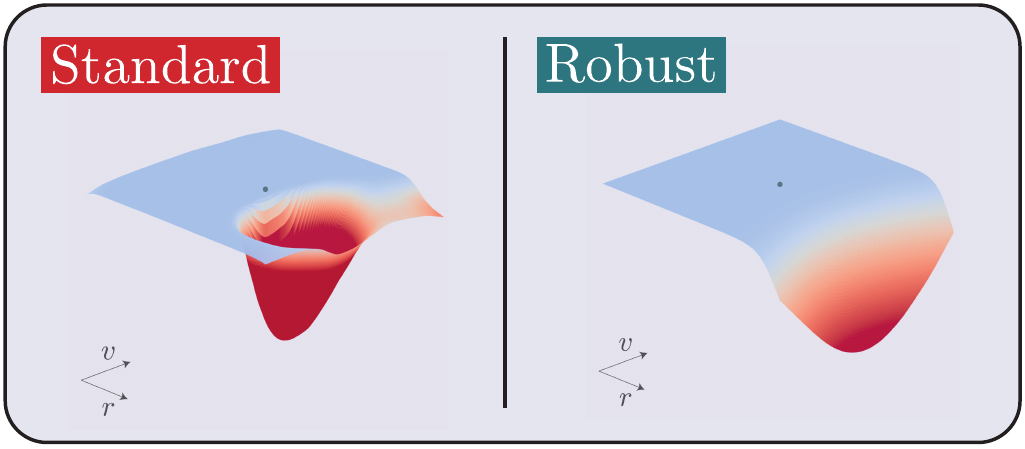}
    \caption{Cross-section of the loss-landscape}
    \label{fig:smooth_boundary}
\end{subfigure}
\caption{Curvature properties of a deep neural network (ResNet-18~\cite{resnet}) trained on an image classification task (CIFAR-10~\cite{krizhevskyLearningMultipleLayers2009}), both using standard training and adversarial training with $\ell_2$ adversarial examples crafted using DeepFool. The network trained using a standard training scheme presents a local loss landscape with a high curvature in certain (adversarial) directions. Meanwhile, the adversarially trained network is much flatter (loss landscape has lower curvature). Image adapted from~\cite{moosavi-dezfooliRobustnessCurvatureRegularization2018} with permission from the authors.}
\label{fig:curvature_profile}

\end{figure}

We define the \textit{local geometry} of a neural network as the geometric properties of the input space of a deep classifier in an $\epsilon$-neighborhood of a data sample $\x$. By construction, most adversarial examples are found in the vicinity of some sample $\x$ and lie close to the decision boundary. Hence, their characteristics are intimately linked to the local geometry of deep classifiers. For example, the success of adversarial attacks based on first-order methods to approximate \eqref{eq:adv_example} demonstrates that deep classifiers are relatively smooth and simple, at least in the vicinity of data samples. This contradicts the common intuition that, because neural networks have a high capacity, they realize arbitrarily complex functions. Indeed, even if popular adversarial attacks methods like FGSM~\cite{goodfellowExplainingHarnessingAdversarial2015}, DeepFool~\cite{moosavi-dezfooliDeepFoolSimpleAccurate2016}, C\&W~\cite{Carlini_Wagner}, or PGD~\cite{madryDeepLearningModels2018} are prone to get trapped in local optima in non-convex settings, their relative success demonstrates that the local geometry of deep neural networks is, in practice, approximately smooth\footnote{Note that ReLU networks can only realize piece-wise linear functions which by definition are non-differentiable in the intersection of two adjacent linear regions. However, in a slightly coarser scale -- bigger than that of the linear regions -- ReLU networks can be approximated by smooth functions.} and free from many irregularities (see Figure~\ref{fig:cross_sections} and Figure~\ref{fig:curvature_profile}).

The aforementioned attacks can, therefore, be used to identify points lying exactly on the decision boundary, i.e., \textit{minimal adversarial examples}. These points inform of important features of the local geometry of a neural network. In particular, the minimal $\ell_2$ adversarial perturbation $\r_2^\star(\x)$ of a sample $\x$ permits to construct a good approximation of the direction that is normal to the decision boundary around $\x$ (see Figure~\ref{fig:diagram_adv}). Surprisingly, these normals are very correlated among different data points, and the set of all $\r^\star_2$ perturbations of a network only spans a low-dimensional subspace of the input space~\cite{tramer2017SpaceTransferable,moosavi-dezfooliRobustnessClassifiersUniversal2018}. Besides, the subspace of normals from different networks is also aligned~\cite{tramer2017SpaceTransferable}. This explains the intriguing observation that adversarial examples transfer well between different architectures, i.e., the adversarial perturbation on a sample $\x$ of a network $f_\th$ is likely to fool also another network $g_{\bm{\phi}}$ trained on the same dataset.

Other important geometric insights come from the study of different perturbations besides the $\r_2^\star(\x)$. In fact, because one can generally fool a network at a given sample in multiple directions~\cite{tramer2017SpaceTransferable,fawziEmpiricalStudyTopology2018,ortiz-jimenezHoldMeTight2020, fawziRobustnessClassifiersAdversarial2016}, analyzing the geometric properties of the boundary in those directions is very informative. Specifically, the cross-sections of the boundary using planes spanned by different types of perturbations (see Figure~\ref{fig:cross_sections}) highlight the flatness of the boundary in some directions, and its high curvature in others. This naturally leads to the investigation of curvature as a central property of the local geometry of a neural network. 

To capture this geometry, it is useful to think in terms of the loss landscape\footnote{For simplicity, we choose to introduce the main notions using the loss landscape formalism, but note that the same findings also apply to the geometry induced by the logits (final layer of the classifier before the softmax operation) of a network.} in the input space induced by the classifier $f_\th$. The loss landscape of a neural network is a function $\ell:\R^D\rightarrow\R$ that maps any input sample $\x$ to
\begin{equation}
    \ell(\x)=\L(\x,y;\th),
\end{equation}
where $y$ is usually taken as the true label of $\x$. Using this function we can approximate the local geometry of a neural network using a second-order Taylor decomposition around $\x$ and write
\begin{equation}
    \ell(\x+\r)\approx \ell(\x)+\r^T\nabla_\x\ell(\x)+\cfrac{1}{2}\,\r^T\nabla_\x^2\ell(\x)\r,
\label{eq:loss_quadratic}
\end{equation}
where $\nabla_\x\ell(\x)$ denotes the gradient of the loss with respect to the input, and $\nabla_\x^2\ell(\x)$ its second-order derivative, or Hessian. Studying the terms in this decomposition for a trained neural network, we can discover that $\r^\star_2(\x)$ and $\nabla_\x\ell(\x)$ are very aligned in most networks. This is mostly due to the fact that the higher order terms in the decomposition are small. Hence, we can say that the network is approximately locally linear. This explains why adversarial attacks that only use the gradient $\nabla_\x\ell(\x)$ are successful in many settings~\cite{goodfellowExplainingHarnessingAdversarial2015}. 

However, it is clear that neural networks are not completely linear. As the eigendecomposition of $\nabla_\x^2\ell(\x)$ demonstrates, there exist a few very curved directions in the loss landscape~\cite{fawziEmpiricalStudyTopology2018, moosavi-dezfooliRobustnessClassifiersUniversal2018} (see Figure~\ref{fig:curvature_profile}).  The predominance of zero eigenvalues in the curvature profile explains the robustness of neural networks to random noise. Indeed, for most random directions the neural networks behave similar to linear classifiers, and hence are robust to random perturbations~\cite{fawziRobustnessClassifiersAdversarial2016}. Meanwhile, the highly curved directions explain the susceptibility of the networks to multiple adversarial attacks~\cite{moosavi-dezfooliRobustnessClassifiersUniversal2018}\footnote{Note that curved decision boundaries are not necessarily easier to attack than flat ones for general classifiers. In fact, how easy it is to fool a data sample depends on the distance to the boundary and not the shape of it. In neural networks, however, it seems that there is a correlation between the curved directions of the loss landscape and the directions in which the neural network can be adversarially attacked.}. 

\begin{figure}[t]
    \centering
    \includegraphics[width=0.9\columnwidth]{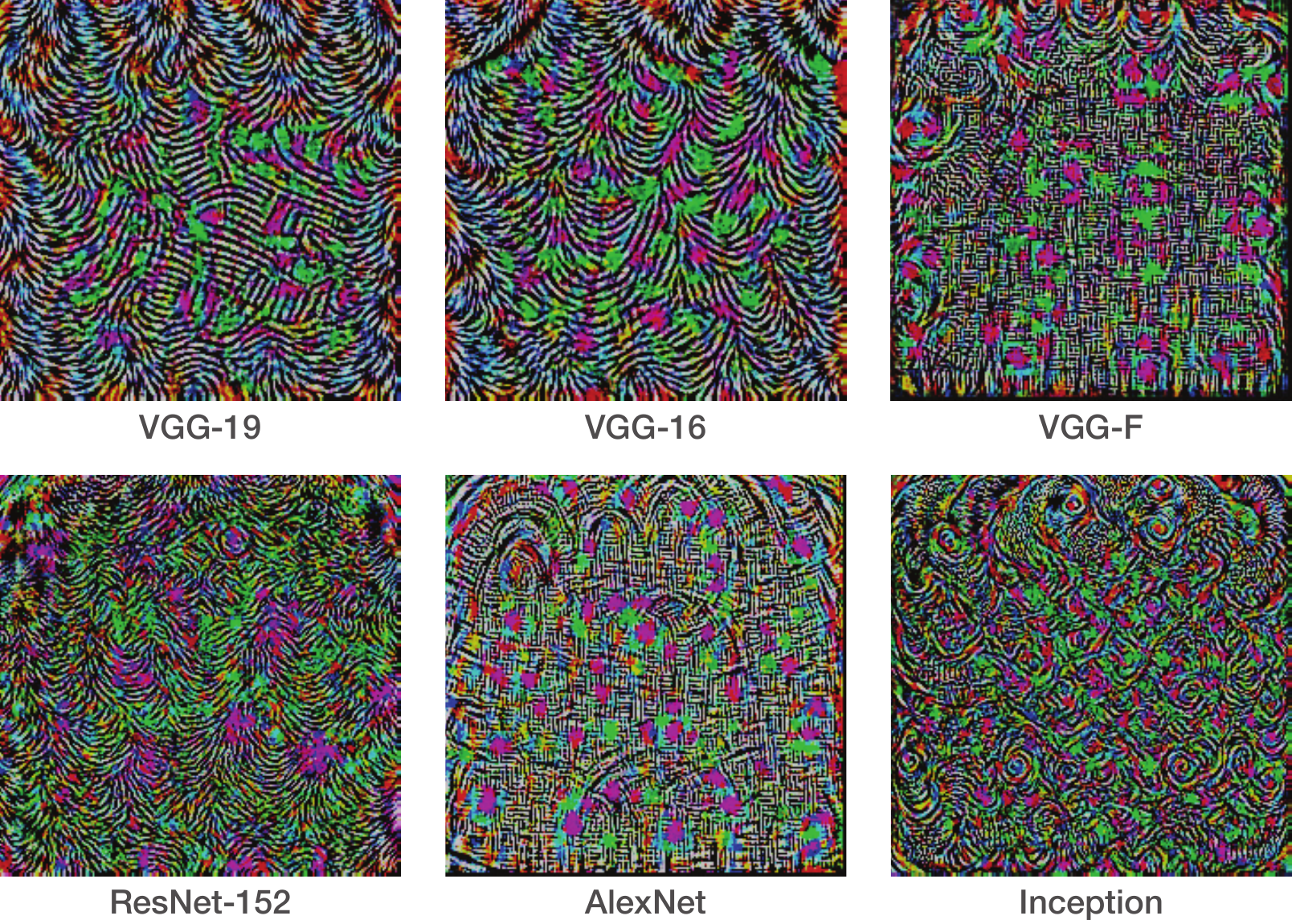}
    \caption{Universal adversarial perturbations (UAPs) computed for different deep neural network architectures. The pixel values are scaled for visibility. When adding a UAP to any image, the classifier is fooled into thinking the sample belongs to an erroneous class. Image taken from~\cite{moosavi-dezfooliUniversalAdversarialPerturbations2017} with permission from the authors.}
    \label{fig:uaps}
\end{figure}

Furthermore, it has been shown~\cite{fawziEmpiricalStudyTopology2018, moosavi-dezfooliRobustnessClassifiersUniversal2018} that the principally curved directions of a deep classifier are also aligned, in the same way that the normals to the decision boundary are correlated among data points. This gives rise to an explanation for another intriguing phenomenon in adversarial robustness: the existence of \emph{universal adversarial perturbations} (UAPs)~\cite{moosavi-dezfooliUniversalAdversarialPerturbations2017}, i.e., constant adversarial perturbations $\r_\text{univ.}$  which, independently of the data point to which they are added, can fool a deep classifier with high probability. That is
\begin{align}
    \r_\text{univ.} =& \arg\min_{\r\in\mathcal{C}} \|\r\|_p\\
    & \text{ s.t. } \mathbb{P}_{(\x,y)\sim\mathcal{D}}\left(f_\th(\x)\neq f_\th(\x+\r)\right)\geq 1-\delta,\nonumber
\end{align}
where $0\leq\delta< 1$ controls the misclassification probability of the perturbation (see Figure~\ref{fig:uaps} for some examples of UAPs for ImageNet~\cite{dengImageNet}). The connection between curvature and UAPs stems from the fact that most of the energy of UAPs is concentrated in the subspace spanned by the shared directions of high curvature. 

\subsection{Why geometry matters}
\label{sec:geometry_applications}

A geometric perspective on adversarial robustness permits to reveal multiple geometric features of deep neural networks, as discussed above. We show, now, how geometric properties can be leveraged to further improve and evaluate the robustness of deep classifiers. In fact, the ``true'' robustness is intrinsically a geometric concept, as it necessarily implies that the distance between the data points and the decision boundaries should be large. This is, achieving robustness implicitly requires to push the decision boundary as far as possible from all data points. This is fundamentally different from the ``false'' sense of robustness, in which a defense method simply protects the model from a specific adversarial attack, or ``hides'' the non-robust features of the model rather than robustifying it~\cite{athalyeObfuscatedGradientsGive2018,carlini2019OnEvaluating,gowal2020uncovering}. This means that a  method should actually have a real impact on the geometry of the classifier in order to achieve ``true'' robustness. 

A better understanding of the ideal geometric properties of a truly robust model is, therefore, key to attain more robust systems. An effective way to investigate these properties is by observing how adversarial training affects the geometric characteristics of the decision boundary. The first -- and perhaps expected -- feature to notice is that adversarially trained networks seem to create boundaries that are further apart from most data samples~\cite{moosavi-dezfooliDeepFoolSimpleAccurate2016}\footnote{For example, this can be easily seen by comparing the norm of the minimal $\ell_2$ adversarial perturbations between the two differently trained models, or by measuring the accuracy of both models on $\epsilon$-constrained adversarial examples.}. However, the most interesting geometric change is the fact that the decision boundaries of adversarially trained networks exhibit a lower mean curvature than that of standard models~\cite{moosavi-dezfooliRobustnessCurvatureRegularization2018} (see Figure~\ref{fig:curvature_profile}). In fact, it seems that this flatness is not just a by-product of adversarial training, but rather a general property of more robust models: One can design computationally efficient defenses that directly regularize the curvature during standard training and achieve comparable robustness to that of adversarial training~\cite{moosavi-dezfooliRobustnessCurvatureRegularization2018,QinLocalLinearization,SinglaFeiziSecondOrder}. 

The lower curvature of robust models can also be exploited to achieve robustness with guarantees, i.e., theoretical certificates that demonstrate a specific level of robustness for a given neural network. Indeed, certifiable adversarial defenses, like randomized smoothing~\cite{CohenCertified, YangRandomizedSmoothing,salman_provably_2020}, also implicitly regularize curvature by averaging the decision of a classifier on randomly perturbed samples. This way, one effectively convolves the loss landscape of a classifier with the probability density function of the perturbation distribution, hence, reducing the mean curvature of the loss landscape and smoothing the input geometry of the classifier.

\begin{figure}[t]
    \centering
    \includegraphics[width=0.9\columnwidth]{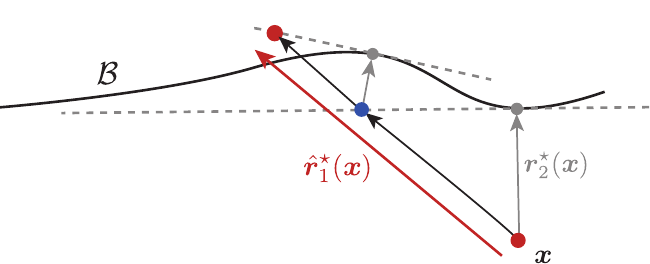}
    \caption{Sketch of how the approximate linearity of the decision boundary around a typical data sample can help to compute other types of adversarial attacks, e.g.,~$\r_1^\star(\x)$. Because the linear approximation of the boundary is fairly accurate, instead of solving \eqref{eq:adv_example} directly one can \emph{iteratively} use the linear approximation of the classifier to solve an easier surrogate problem and efficiently find an approximate solution to the original problem, e.g.,~$\hat{\r}_1^\star(\x)$ as in~\cite{ModasSparseFool}.}
    \label{fig:sparsefool}
\end{figure}

A proper understanding of the geometric properties of neural networks can also be exploited to design efficient adversarial attacks, or to effectively evaluate robustness in different settings. Recall that, in most settings of interest, solving \eqref{eq:adv_example} might be rather challenging, especially when working with metrics beyond the standard $\ell_2$ and $\ell_\infty$ distances. In such cases, the low mean curvature of the decision boundaries in the vicinity of data samples~\cite{moosavi-dezfooliRobustnessClassifiersUniversal2018, fawziEmpiricalStudyTopology2018,jetleyFriendsTheseWho2018} can be exploited to design more efficient adversarial attacks. In fact, a useful relaxation of \eqref{eq:adv_example} is to linearize the decision boundary in the vicinity of the data point $\x$. Finding an adversarial attack becomes quite effective in this linearly approximated problem. Because of the low mean curvature, the linearization is usually close to the real problem, hence enabling the adversarial attacks to successfully converge to a solution. This ``local flatness'' property has been very important in the design of computationally efficient adversarial attacks in challenging settings like the $\ell_0$ or $\ell_1$~\cite{ModasSparseFool} metrics (see Figure~\ref{fig:sparsefool}). Similarly, it has also been exploited to craft black-box attacks~\cite{LiuQFool,RahmatiGeoDA}, in the challenging context in which the attacker has no access to the weights of the target classifier, except from the output label or the class probabilities.

\subsection{Discussion}\label{sec:discussion_adv_examples}

At this stage, it is important to highlight that the geometric insights discussed so far, have been mainly studied in the context of computer vision applications and image data. In principle, most of these observations should also translate to other domains like natural language processing (NLP) or tabular data. However, their relevance would probably hinge on whether the $\ell_p$ metric defines a reasonable topology for some representation in these domains, e.g., the embedding space of a language model. In general, more research is necessary to understand and explain the general geometric properties of neural networks trained on other types of data and with different architectures.

Right after their discovery in deep neural networks, adversarial examples were soon labeled as ``holes'' of deep learning systems, posing serious security threats to their real-world applications~\cite{modas_toward_2020,kurakin_adversarial_2017,EykholtRobustPhysical,ilyas18aBBox,Guo19BBox,bhambri2019survey}. This exposed vulnerability quickly led to an ``arms race'' between adversarial attacks and defenses whose main goal was to come up with the most secure model. Nevertheless, the existence of adversarial examples also shows that deep neural networks do not base their decisions in the same abstractions that humans use. This questions the reliability and trustworthiness of these systems. Many properties of adversarial examples actually suggest that their existence is tightly linked to the way neural networks learn. Interestingly, this also means that adversarial examples may reveal important information about the inner workings of deep neural networks. The main evidence supporting this idea relies on the following observations:
\begin{itemize}
    \item Adversarial examples are pervasive across most applications of deep learning.
    \item They are easy to compute, demonstrating the simplicity of the decision boundaries.
    \item They transfer across networks trained on the same dataset.
    \item There seem to be very strong structural similarities among adversarial examples for different data points, e.g., we can compute UAPs.
\end{itemize}
All these properties hint towards the fact that adversarial examples are a reflection of the decision process of a neural network, and not just loopholes in its functioning. 

The main conceptual challenge with this perspective is that it requires to acknowledge the fact that there might be multiple functions -- or classifiers -- that can effectively exploit the information in the training data and generalize to the unseen test distribution. That is, the solution to \eqref{eq:accuracy} might not be unique. As a result, not only there might be multiple solutions that can perfectly fit the training data, in the sense of \eqref{eq:erm}, but also there might be more than one solution that can generalize to the test data in \eqref{eq:accuracy}. Informally speaking, one can see humans and neural networks as two classifiers with good generalization capacity that leverage different information. Indeed, neural networks have good test accuracy, but the existence of adversarial examples is a strong evidence that the way neural networks generalize is by exploiting different features of the data than those humans use. In the next section, we will delve into this connection and review the main works supporting this idea.

\section{Understanding deep learning through the lens of adversarial robustness}\label{sec:understanding_dl}

In this section, we review the main works in the intersection of adversarial robustness and deep learning theory, presenting the main established results, but also highlighting the main open research questions. Indeed, as discussed in Section~\ref{sec:geometric_view}, the existence of adversarial examples manifests more than just a security threat. Analyzing the adversarial robustness properties of deep neural networks has actually revealed many of their characteristics, so that adversarial robustness can be carefully exploited to enhance our understanding of deep learning systems. 

This section is divided in two main axes. We first review the connection between adversarial robustness and generalization (Section~\ref{subsec:adversarial_robustness_and_generalization}), and second, we show how adversarial robustness informs about the learning dynamics of neural networks (Section~\ref{subsec:dynamics_learning}).

\subsection{Adversarial robustness and generalization}
\label{subsec:adversarial_robustness_and_generalization}

The theoretical connection between adversarial robustness and generalization can be summarized in the following questions: How is it possible that neural networks generalize so well, when their outputs are so sensitive to perturbations? Or, why do most adversarial defenses, such as adversarial training, hurt performance on the clean test set? Such questions have puzzled machine learning researchers, which do not seem to agree on the right answer. Some researchers believe that adversarial robustness is at odds with accuracy and that there must be a fundamental trade-off between these two notions~\cite{Fawzi_Analysis_MachineLearning, tsiprasRobustnessMayBe2018,fawziAdversarialVulnerabilityAny2018,RaghunathanUnderstanding}. Meanwhile, others believe that in some cases robustness is positively correlated with performance~\cite{gilmerAdversarialSpheres2018}. While one can find empirical evidence in favor of both arguments, a definite theoretical resolution remains to be found. Despite the current debate, however, there is a general consensus about the tight connection between the adversarial robustness and the generalization phenomena. We will discuss this connection below from three different perspectives.

\subsubsection{The distribution of adversarial perturbations can predict generalization}

A strong case in favour of the connection between adversarial robustness and generalization can be found in \cite{jiangPredictingGeneralizationGaph2019,WerpachowskiDetectingOverfitting}, where researchers investigate a simple question: Is it possible to predict the test accuracy of a neural network given only access to training data? This question becomes particularly important when the number of available test samples are not enough to reliably measure the performance of a trained model. Towards measuring the generalization performance based on the training data, they propose to use the distribution of distances to the decision boundary of a neural network, measured at different training points and at different layers, to predict the generalization gap of the model. Their main claim is that, as in support vector machines (SVMs)~\cite{cortes_vapnik_1995} where margin controls the generalization capacity of the model, the distribution of $\|\r_2^\star(\x)\|_2$ does the same on neural networks\footnote{Note that the notion of minimal adversarial perturbations, i.e., $\r_2^\star(\x)$, in a neural network is strongly related to the concept of margin. Indeed, $\|\r_2^\star(\x)\|_2$ measures the distance to the boundary for a sample $\x$.}. They confirm this by training a logistic model on the distribution of the first moments of these margins, and show that the generalization gap of these networks can be estimated with good precision. This is the same as saying that the local geometry of a deep classifier around different training samples determines its generalization ability. Parallel studies have also confirmed that a smaller sensitivity to input perturbations correlates well with better generalization~\cite{novak2018sensitivity}. In general, it has been widely observed that the distance to the decision boundary of different data samples is an important quantity for generalization.

\subsubsection{Adversarial examples are features}\label{subsec:robust_features}

Traditionally, deep learning is introduced as a machine learning technique that does not require to use handcrafted features~\cite{lecun_bengio_hinton_2015}. Instead, deep neural networks are supposed to find ``good'' features of the training data, for a given learning task. It thus means that the key to the success of deep learning is the choice of features exploited by a neural network. But, what are these features exactly? Could it be that the adversarial examples themselves are a reflection of these features?

The transferability of adversarial examples and the existence of UAPs can partly answer these questions. In fact, authors in~\cite{jetleyFriendsTheseWho2018} showed that, if one projects the test dataset onto the principal components of a collection of $\r_2^\star(\x)$ adversarial examples -- which happen to be very aligned with the shared curved directions as well~\cite{fawziEmpiricalStudyTopology2018} --, the accuracy of a neural network trained on this dataset barely changes. This means that the same directions that can be used to attack a deep network are the ones that the network exploits to classify the data. Furthermore, the fact that this simple projection works, combined with the shared geometric properties of the boundary in the vicinity of data points -- low-rank structure of minimal adversarial perturbations at different data points, and UAPs --, suggests that deep classifiers are mainly exploiting simple and brittle features of the datasets, i.e., non-robust features that are aligned with adversarial perturbations. In other words, these are features that seem to correlate well with the output labels, but can be easily perturbed to fool a classifier.

\begin{figure}[t]
    \centering
    \includegraphics[width=\columnwidth]{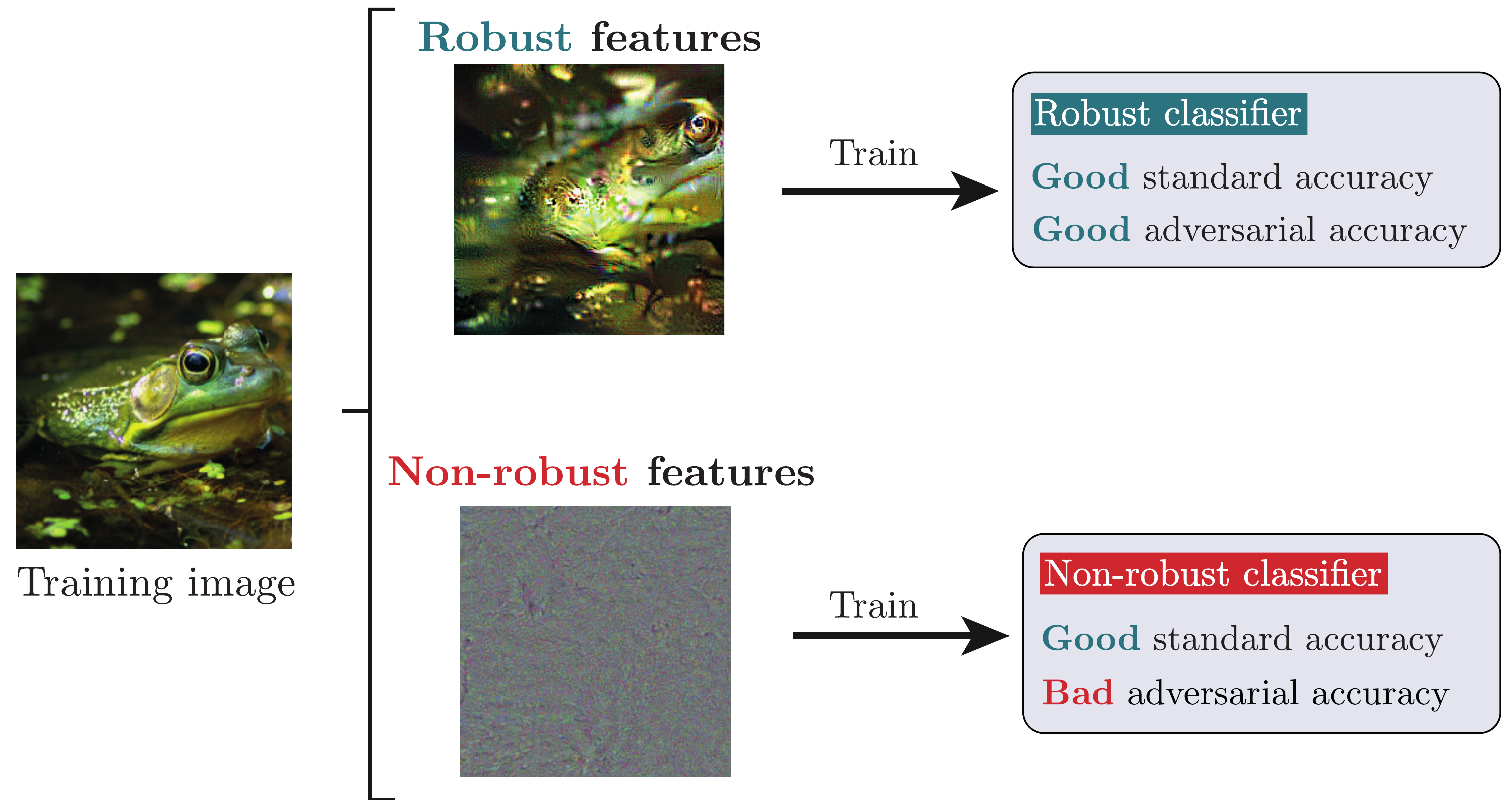}
    \caption{Diagram explaining how an image can be decomposed in robust and non-robust features. The non-robust features are sufficient to achieve good standard generalization, but result in brittle classifiers which are vulnerable to small adversarial attacks. On the other hand, the robust features of a dataset are also good enough to achieve good generalization, but instead the resulting classifiers are more robust to adversarial attacks. While, the non-robust features resemble the adversarial perturbations of standard models, the robust ones are semantically more human-aligned. Image adapted from~\cite{ilyasAdversarialExamplesAre2019} with permission from the authors.}
    \label{fig:non-robust}
\end{figure}

The above non-robust feature hypothesis is further supported by additional recent evidence. In particular, the authors in~\cite{ilyasAdversarialExamplesAre2019}  showed that it is possible to construct a dataset consisting only of the non-robust features of a large-scale vision dataset. To do so, they took a restricted subset of a classical image database called ImageNet~\cite{dengImageNet}, and computed adversarial examples $\r_2^\epsilon(\x^{(i)})$ on all its training samples $\{\x^{(i)}\}_{i=1}^N$. They assigned the adversarial labels $f_\th(\x^{(i)}+\r_2^\epsilon(\x^{(i)}))$ to these samples, which were eventually used to train another regular deep network. Note that, in this modified dataset, the labels do not correlate with our human perception but include some imperceptible features that can be used to fool a deep classifier. Still, the newly trained classifier showed non-trivial accuracy on the original unmodified test set, albeit with low adversarial robustness (see Figure~\ref{fig:non-robust}). This experiment corroborates that there exist non-robust ways to fit the training data, which can still lead to good generalization. Indeed, the only way that the network trained on the adversarial samples can generalize to the unmodified test set is by exploiting the features introduced by the adversarial perturbations themselves, i.e., the non-robust features.

Note here that the definition of robust and non-robust features has remained vague in the adversarial robustness literature to this date. In particular, the works above tend to loosely refer to the difference between robust and non-robust features as an abstract proxy to refer to the idea that standard and robust classifiers have different ways to separate the data. The language of causality~\cite{ElementsOfCausalInference}, however, provides a rigorous framework to address this problem. In particular, the field of causal statistics provides an established formal framework to distinguish between variables  that have a spurious association with the target, and features that are causally relevant. In fact, some recent works have already explored some of the connections between distributional robustness\footnote{Distributional robustness is a field of statistics concerned with obtaining worst-case estimators within some family of distributions of the training data. This differs slightly from the notion of adversarial robustness discussed in this review for which the robustness constraints are defined in terms of perturbations of single samples. and causality~\cite{CausalityDistributionalRobustness, heinzedeml2019conditional}. Specifically, they provide a causal framework to define robustness to confounding variables and to propose some algorithmic estimators. Other works have suggested using counterfactually-augmented data\footnote{Counterfactual examples are conceptually equivalent to adversarial examples. Traditionally, the main difference lies in the distance metric used to define the allowable perturbation. See Section~\ref{sec:interpretability} for a more detailed discussion.} to drive the model towards learning causal paths and avoiding spurious correlations~\cite{Kaushik2020Learning}. Counterfactual data augmentation is conceptually very similar to adversarial training, with the difference that the data perturbations need to be manually prescribed by a human annotator. Despite this preliminary effort, the connection between adversarial robustness and causality is still largely unexplored. This is, therefore, a promising path for cross-fertilization.
}

Showing that adversarial examples are tightly connected to the generalizable features of the training set is an important landmark for the field of adversarial robustness. It not only gives an explanation why adversarial examples exist in the first place, but also, it provides a constructive way to improve the feature representations of neural networks. Indeed, the fact that more robust classifiers have features which align better with our human perception is one of the main enablers of the wide range of applications of adversarial robustness beyond security (more on this in Section~\ref{sec:applications_robustness}).
 
\subsubsection{Adversarial robustness and accuracy trade-off}
\label{subsubsec:adv_robustness_acc_tradeoff}
Following the above reasoning, one can say that the role of adversarial training -- as one of the best methods to improve robustness of deep networks -- is to filter out, or purify~\cite{allen-zhu_feature_2020}, the training set from non-robust features. Hence, it forces the network to learn robust representations instead. This perspective gives a good argument in favour of the hypothesis that adversarial robustness might be at odds with accuracy~\cite{Fawzi_Analysis_MachineLearning, tsiprasRobustnessMayBe2018,fawziAdversarialVulnerabilityAny2018}. Furthermore, it justifies the empirical observation that one needs more data to generalize when using adversarial training~\cite{schmidtAdversariallyRobustGeneralization2018}. Indeed, if the non-robust features are useful to achieve good standard generalization, then it is reasonable that removing them from the training set might hurt clean accuracy, or that one needs more data to generalize.

On the other hand, it seems like humans are not so susceptible to adversarial perturbations (at least the ones constructed for deep neural networks), but can still achieve good accuracy on recognition tasks, suggesting that humans do not use non-robust features~\cite{serre_deep_2019}. For these reasons, some researchers argue that there has to be some way in which exploiting only robust features of a dataset can achieve good, or even better, clean accuracy~\cite{Xie_Improve, Zhu2020FreeLB, TangOnlineAugment}. In fact, some theoretical results indicate that there exist at least some synthetic distributions, in which adversarial robustness and accuracy are positively correlated~\cite{gilmerAdversarialSpheres2018,dohmatobNoFreeLunch}. In these distributions, adversarial examples are a naturally occurring phenomenon caused by the high-dimensionality of the data representation, and can be proved to be connected to the amount of test error of any classifier. This means that the only way to improve the robustness of a classifier trained on these distributions is to improve the performance on the unperturbed test set, and vice-versa.

Whether there is a fundamental trade-off between adversarial robustness and accuracy, or whether the current techniques are sufficient to achieve the optimal performance in real data is still an open research debate. In the meantime, recent works are showing that adding more data -- even unlabeled -- into the adversarial training process can improve the robustness of deep classifiers and decrease the accuracy gap~\cite{AlayracAreLabelsRequired, CarmonUnlabeledAT, RaghunathanUnderstanding, zhai2019unlabeled, najafi2020incomplete}. Nevertheless, some recent theoretical studies have shown that there are at least some distributions for which adding more data to adversarial training can hurt clean generalization~\cite{ChenMoreDataCanExpandGap}. The regime under which this happens, however, depends on the strength of the adversary and seems to predominantly appear when requiring $\epsilon$ in \eqref{eq:epsilon_constraint} to be very large. More research should therefore be devoted to understand if this phenomenon can also affect data distributions used in practice. 

\subsection{Dynamics of learning}\label{subsec:dynamics_learning}

Understanding how a neural network evolves during training is a complicated yet crucial problem that has lately attracted a lot of attention. Deep neural networks tend to comprise millions of parameters, and their training dynamics across the optimization landscape are hard to analyze. Nonetheless, being able to explain what and how neural networks learn is fundamental to understand their behaviour. 

Recently, some researchers have started to use adversarial examples to shed light on these complicated dynamics. In particular, the use of adversarial proxies to track the evolution of the geometry of these systems, and the analysis of adversarial training as an alternative training algorithm, can provide important insights towards explaining the dynamics of deep learning. In this section, we review the main results in this emerging field.

\subsubsection{Adversarial examples can be used to track the evolution of a neural network during training} The intimate link between adversarial examples and the features exploited by deep models provides a novel way to study deep neural networks based on their adversarial properties. In this sense, an efficient way to analyze the transformation of a neural network during training is to track the evolution of its adversarial perturbations~\cite{yinFourierPerspectiveModel2019,ortiz-jimenezHoldMeTight2020}. Such an analysis gives insights about the state of the local geometry of a neural network in the input space and can inform about many of its properties.

One such property is the inductive bias of deep learning towards invariance. The inductive bias of a learning algorithm refers to the preference of a model to select one among the multiple possible solutions that fit the training data~\cite{mitchellNeedBiasesLearning, battagliaRelationalInductiveBiases2018}. Invariance relates as well to the tendency of a neural network to ignore non-discriminative features of the training data and thus assign the same label to large regions of the input space. Note that this is a generally expected property of any learning algorithm, but, in deep learning, the existence of adversarial examples contends this idea, as it highlights the fact that the decision boundary of a neural network is always very close to any data sample.

Nevertheless, the presence of this inductive bias in deep neural networks has recently been confirmed~\cite{ortiz-jimenezHoldMeTight2020}. To show this, the researchers manipulated the features of the most common computer vision datasets, altering the data in distinct frequency bands. They then looked at the impact of this data manipulation on the decision boundary of a neural network, when trained on such modified data. 

To inspect the decision boundary, they estimated the minimal adversarial perturbations $\r_{2,\mathcal{S}_i}^\star(\x)$ of a neural network in a set of orthogonal subspaces $\mathcal{S}_i\subseteq\R^D$, i.e.,
\begin{equation}
    \begin{split}
        \minimize{\r\in\R^D}&\quad \|\r\|_2\\
        \text{subject to }&\quad f_\th(\x+\r)\neq f_\th(\x)\\
        &\quad \r\in\mathcal{S}_i,
    \end{split}
\label{eq:minimal_subspace}
\end{equation}
whose norm $\|\r^\star_{2,\mathcal{S}_i}(\x)\|$ gives a good estimate of the distance of the decision boundary within that subspace. The key to this experimental setup is to select the collection of inspecting subspaces $\{\mathcal{S}_i\}_{i=1}^S$ such that the interventions done to the training data lie on different subspaces. Doing this, researchers discovered that neural networks can only create decision boundaries along those subspaces where they identify discriminative features. And, as a result, the distance to the boundary outside these subspaces is very large, which relates to invariance in these subspaces\footnote{The capacity of neural networks to distinguish between discriminative (with relevant information for classification) and non-discriminative (containing noise, or information irrelevant for classification) subspaces has recently been shown theoretically to be one of the main advantages of deep learning over kernel methods~\cite{ghorbani2020neural}.}.

\begin{figure}[t]
\begin{subfigure}[b]{0.49\columnwidth}
    \centering
    \includegraphics[width=\columnwidth]{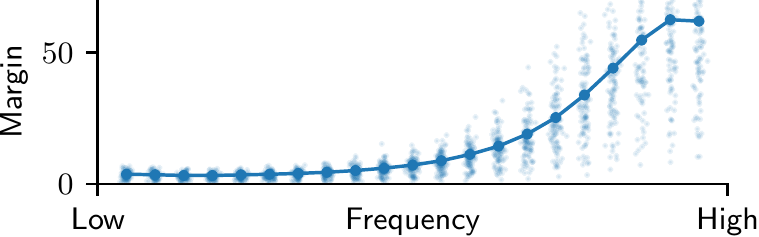}
    \caption{LeNet (MNIST)}
\end{subfigure}
\hfill
\begin{subfigure}[b]{0.49\columnwidth}
    \centering
    \includegraphics[width=\columnwidth]{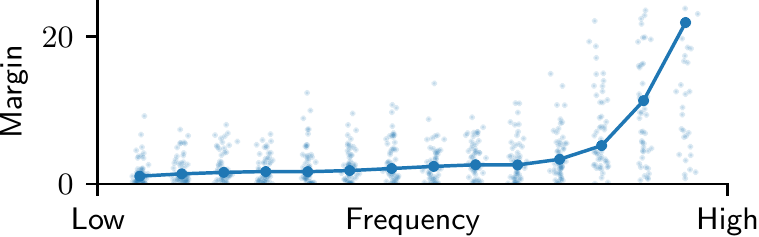}
    \caption{ResNet-50 (ImageNet)}
\end{subfigure}
\label{fig:hold_me_tight}
\caption{Margin distribution of test samples in subspaces taken from the diagonal of the DCT (low to high frequencies). The thick line indicates the median values of the margin, and the shaded points represent its distribution. The low margin values (low frequencies) correspond to subspaces where the network identifies discriminative features, while the high margin to subspaces the network is invariant to. Image taken from~\cite{ortiz-jimenezHoldMeTight2020} with permission from the authors.}
\end{figure}

This provides a direct way to study if some information in the training set is exploited for classification, by measuring the average margin of a network in different subspaces. For example, this test has been used to explain why convolutional neural networks are more vulnerable to low-frequency perturbations than high-frequency ones~\cite{yinFourierPerspectiveModel2019,sharmaEffectivenessLowFrequency,tsuzukuStructuralSensitivityDeep2019,RahmatiGeoDA}, or why the existence of redundant features in a dataset -- information in the data which might be exploited to create decision boundaries, but is not necessary for generalization -- can hurt the robustness of a classifier to small shifts of its test distribution~\cite{OrtizJimenezRedundantFeatures}. Furthermore, it gives a new angle on the problem of catastrophic forgetting~\cite{mccloskeyCatastrophicInterferenceConnectionist1989} -- when a trained neural network is fine-tuned on new data it tends to lose its accuracy on the previous distribution -- as it demonstrates that the decision boundaries of a neural network can only exist for as long as the classifier is trained with the discriminative features that support them.

\subsubsection{Understanding adversarial training} Adversarial training connects deep learning to a long tradition of robust optimization research in other fields. Before the advent of deep learning, robust and distributionally robust optimization were active research areas in the context of SVMs~\cite{XuRobustSVM}. For these classifiers, the objective \eqref{eq:adv_training} can be written as a convex optimization problem for certain constraint sets $\mathcal{C}$. This makes the task of obtaining robust SVM classifiers amenable to standard convex optimization techniques, which guarantees the convergence to a correct solution~\cite{caramanisRobustOpt,Lanckriet,BhattacharyyaRobustClassification}.

The simplicity of the convex SVM case contrasts with the optimization complexity of deep learning. Indeed, for deep neural networks, the objective \eqref{eq:adv_training} is non-convex and non-concave, and guaranteeing an adequate solution to this problem is not trivial. Nevertheless, in practice, adversarial training, i.e., the replacement/augmentation of training data with adversarial examples crafted during training, seems to be effective to solve \eqref{eq:adv_training} for state-of-the-art image classifiers~\cite{madryDeepLearningModels2018}. Indeed, adversarially trained networks converge to solutions which are more robust than those of standard training, and exhibit different geometric properties~\cite{moosavi-dezfooliRobustnessCurvatureRegularization2018} (see Section~\ref{sec:geometry_applications}). 

\begin{figure}
    \centering
    \includegraphics[width=\columnwidth]{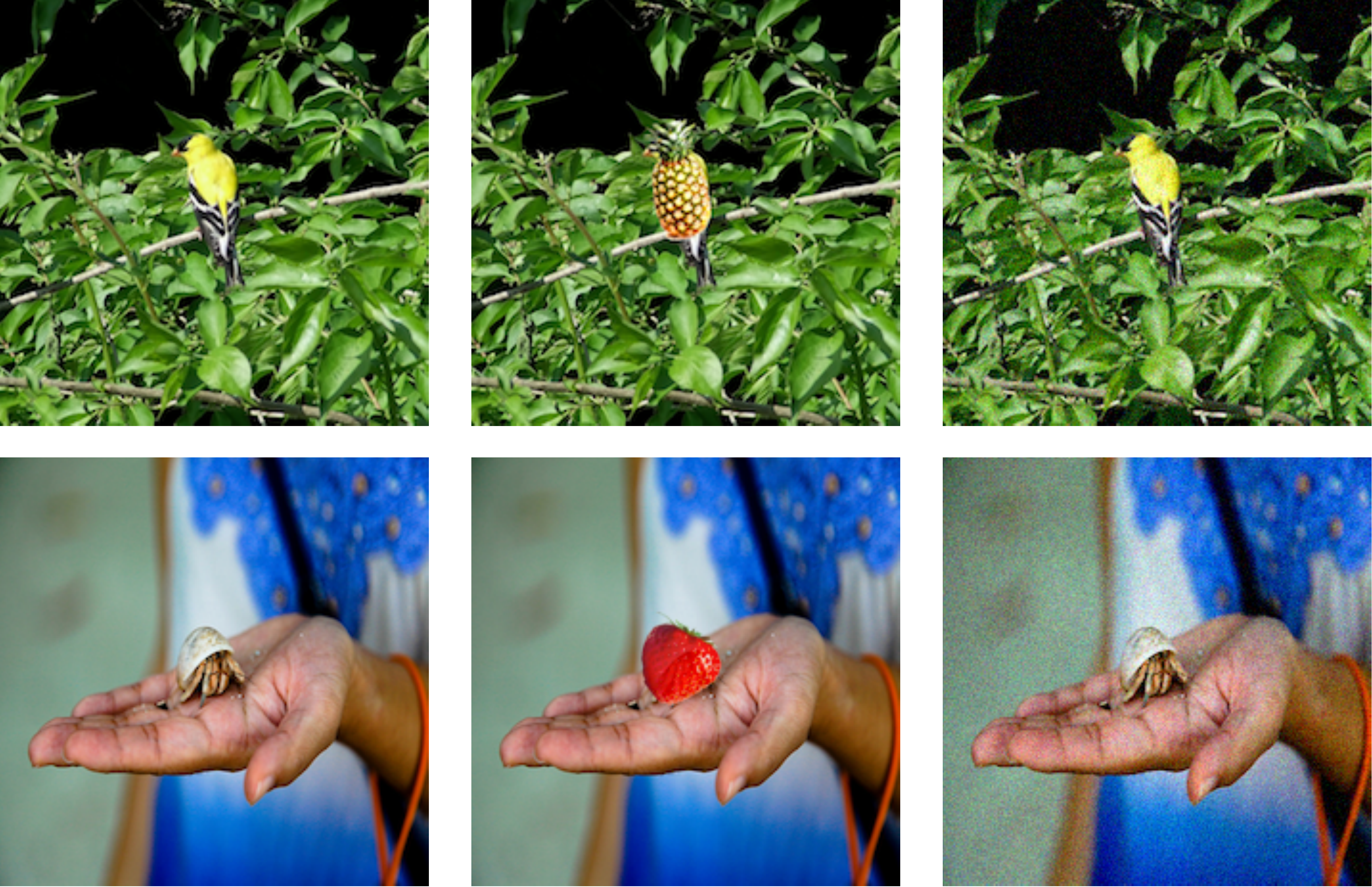}
    \caption{A collection of images in which an overly-robust classifier trained on natural images would necessarily predict the same label. In the middle image, the image on the left has been modified with a very perceptible perturbation of small $\ell_2$-norm which semantically changes the label. However, because the classifier is robust in the $\ell_2$ sense it necessarily needs to label it with the same class as the left image. Hence, it makes a non-human-aligned mistake. For comparison, the image to the right also shows a modification of the image on the left by an $\ell_2$-bounded perturbation with the same norm as the image in the middle. The image on the right is classified according to our human intuition. Image taken from~\cite{tramer_Invariance} with permission from the authors.}
    \label{fig:excessive_invariance}
\end{figure}

The dynamics of adversarial training can reveal important insights of deep learning systems. This training scheme only differs from standard training in that it slightly perturbs the training samples during optimization. However, these small changes can utterly change the geometry of the classifier far beyond the small $\epsilon$-ball that constrains the perturbations~\cite{ortiz-jimenezHoldMeTight2020}. In fact, adversarially trained networks tend to show an even greater invariance than normally trained ones. An excessive invariance that has recently been argued to be prejudicial for classification as it can decrease the alignment between our human perception and the network's decisions~\cite{JacobsenExcessiveInvariance,tramer_Invariance}. This happens because the invariance bias of neural networks alongside their sensitivity to small changes in their training data can force the networks to latch onto \emph{overly-robust} features of the training set which are not human-aligned (see Figure~\ref{fig:excessive_invariance}). This idea seemingly contradicts the general prescription that ``adversarially robust networks are more human-aligned''~\cite{ilyasAdversarialExamplesAre2019}, as it seems that there might also exist some very robust features which escape our human intuition. This is especially true if the threat model is not aligned with our human intuitions, although models robust to $\ell_p$ adversarial perturbations still seem to be more human-aligned than non-robust ones. In general, more research is necessary to understand these trade-offs. It is clear, however, that a better alignment of our formulation of adversarial robustness in terms of metrics that better represent our human perception~\cite{zhou_humans_2019}, would avoid over-optimizing a possibly wrong robustness objective.

Another important difference of some adversarial training schemes with regular training is their natural tendency to overfitting~\cite{RiceRobustOverfitting}. In the deep learning community it is a common belief that neural networks are relatively immune to overfitting~\cite{zhangUnderstandingDeepLearning2016}. Indeed, the train and the test losses of a neural network during training are clearly decreasing. Nevertheless, with adversarial training, it has recently been shown that the best robustness in the test set is consistently found at the middle stages of training~\cite{RiceRobustOverfitting}. This confirms that adversarially trained neural networks have a tendency to overfit. This phenomenon is known as \emph{robust overfitting}, and it has recently been shown to be one reason of some of the reported differences between different adversarial defenses. In fact, the state-of-the-art adversarial training technique uses early-stopping to obtain the best robustness results~\cite{gowal2020uncovering}.

Related to this, another phenomenon that has complicated the evaluation of different adversarial defenses is the complex behaviour of adversarial training when trained using different adversarial attacks. In this sense, it seems that, due to the high sensitivity of neural networks to small changes of their training data, training with different values of $\epsilon$ or using different adversarial attacks to approximate \eqref{eq:adv_example} can change the geometry of the final classifier~\cite{WongFastAT, AndriushchenkoUnderstandingFast}. For example, using FGSM~\cite{goodfellowExplainingHarnessingAdversarial2015} or PGD~\cite{madryDeepLearningModels2018}, a non-iterative and iterative adversarial attack, respectively, to robustify a network against small $\r^\epsilon_p(\bm{x})$ perturbations, is equally effective. However, when using a slightly larger $\epsilon$ for training, the neural networks experience a strong phase transition in terms of the solution found by FGSM-adversarial training and become again vulnerable to adversarial attacks. This does not seem to happen with PGD-adversarial training. This phenomenon is known as \emph{catastrophic overfitting} and it seems to be a consequence of the natural tendency of neural networks to misalign $\r(\x)$ and $\nabla_\x\ell(\x)$ to counteract the effect of the FGSM perturbations crafted only using $\nabla_\x\ell(\x)$~\cite{AndriushchenkoUnderstandingFast}.

To this date, we still lack a precise description of the role of the attacks used in adversarial training. This has lead to an array of common pitfalls in the evaluation of robustness~\cite{carlini2019OnEvaluating, CroceReliable}, but also it might be preventing us from using models with useful properties outside of the commonly accepted robust regimes, e.g., with a specific value of $\epsilon$ (see Section~\ref{sec:applications_robustness}). Understanding the wide range of solutions that can be reached using different types of adversarial training is an interesting open research question for future studies, as it will shed light on how the dynamics of learning affects the behaviour of deep neural networks.

\subsubsection{Importance of regularization}

After the collection of training data, a common deep learning workflow consists in (i) choosing a neural network architecture, (ii) designating an optimizer, (iii) setting some regularization parameters, and (iv) training. If the performance on the validation set is not good enough then one repeats this process with a different set of parameters~\cite{hyperparam_opt}. For deep neural networks, our understanding of the role of any of these steps in the final outcome is rather small, but it is particularly narrow in the case of the regularization. This is in contrast with the classical statistical learning theory, where proper regularization techniques are crucial in training high performance classifiers.

Nevertheless, the study of adversarial robustness has shown that regularization indeed plays an important role in deep learning too, as it can modify the robustness properties of a neural network. In this sense, methods that proposed to regularize curvature~\cite{moosavi-dezfooliRobustnessCurvatureRegularization2018,QinLocalLinearization,SinglaFeiziSecondOrder}, smoothness~\cite{lassance_laplacian_2018,salman_provably_2020}, or gradient alignment~\cite{AndriushchenkoUnderstandingFast} in the input space, have shown that carefully designed regularizers can drive training towards solutions with specific geometric properties in the input space, and push the decision boundary away from the data samples. For this reason, adding explicit regularization to the training objective has been, recently, a great source of inspiration for more efficient adversarial defense schemes. The importance of regularization is, however, not limited to adversarial robustness. Many works have also exploited the ability to modify the geometry of a neural network through explicit regularization, with applications such as improving the stability~\cite{YangInvarianceRegularization,ZhengImrpovingRobustness} or interpretability~\cite{rossImprovingAdversarialRobustness,SalvingRight4Right} of deep learning models.

\section{Applications of adversarial robustness in machine learning}\label{sec:applications_robustness}

As discussed in the previous section, analyzing the adversarial robustness of deep neural networks can shed light on some of their characteristics and properties, hence contributing to our general understanding of deep learning. Adversarial robustness is not only relevant to security or theoretical understanding of deep networks, but it has had a significant impact on many other fields of machine learning, such as anomaly detection~\cite{liang2018enhancing,GoyalDROC}, privacy~\cite{attriguard,songMembership,LecuyerCertified}, or fairness~\cite{Yurochkin2020Training,edwards2015censoring,GargCounterfactual,ZhangMitigating,madras18a,fairness_through_robustness}. We focus in this section on those topics that have attracted the most attention in recent years, and that align better with the perspectives introduced above. These are, interpretability (Section~\ref{sec:interpretability}), transfer learning (Section~\ref{sec:transfer}), and robustness to distribution shifts or out-of-distribution generalization (Section~\ref{sec:common_corruptions}).

\subsection{Interpretability of deep neural networks}\label{sec:interpretability}
Because of the black-box nature of deep learning, it is practically hard to understand how deep networks make their decisions. This can be problematic in sensitive applications like medical diagnosis, or for the deployment of machine learning solutions where biased decisions can have a heavy toll on the lives of discriminated groups. Interpretability, which encapsulates techniques that try to understand and interpret the decisions of neural networks, has therefore attracted a lot of attention in recent years~\cite{Erhan2009, rumelhart1986, RibeiroLIME, Bach2015LRP}.

Interpretability and adversarial robustness are heavily related, not only because both fields make an extensive use of geometry, but also because the recently discovered connection between adversarial examples and features has opened the door to a new framework for interpretability. In what follows, we review some of the main points of connection between the two fields.

\subsubsection{Interpretability and adversarial robustness}

\begin{figure}
    \centering
    \includegraphics[width=\columnwidth]{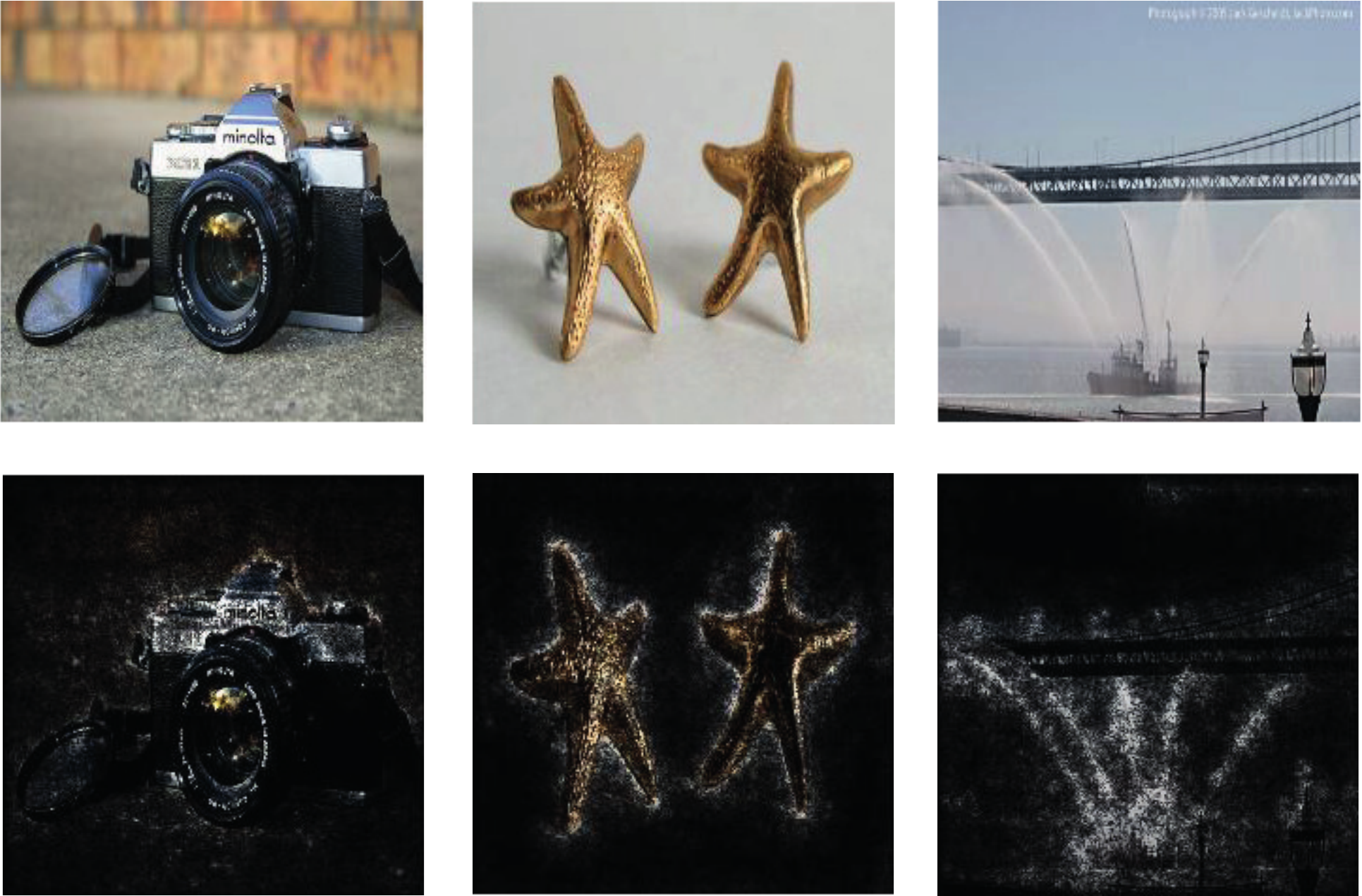}
    \caption{Examples of saliency maps obtained using Integrated Gradients~\cite{sundararajanIntegrtedGradients}. The top row shows the original image, while the bottom row shows the corresponding saliency map. The saliency map highlights the contribution (importance) of each pixel into the network's prediction, and one can notice how the important pixels reflect distinctive image features. Image taken from~\cite{sundararajanIntegrtedGradients} with permission from the authors.}
    \label{fig:saliency}
\end{figure}

In computer vision tasks, a common theme of interpretability methods is to provide interpretations that can be visually inspected directly by humans. These methods are generally known as feature attribution methods~\cite{simonyan2013, AllConvNet, selvarajuGRAD-CAM,sundararajanIntegrtedGradients,Rebuffi_2020_CVPR,FullGrad} and their output is known as saliency maps. Namely, visual representations that highlight the features (i.e.,~pixels) in an image, which are most relevant for a neural network's decision (see~Figure~\ref{fig:saliency}).

The simplest approach to generate saliency maps consists in exploiting the input gradient, $\nabla_\x\ell(\x)$~\cite{simonyan2013}. In this sense, the relative magnitude of each entry in $\nabla_\x\ell(\x)$ can be seen as the importance assigned by the network to a specific pixel $[\x]_i$ for the class $y$. Recall from the discussion over \eqref{eq:loss_quadratic} that $\nabla_\x\ell(\x)$ is highly correlated with $\r_2^\star(\x)$ in most deep classifiers, and that this is precisely the quantity that most adversarial attacks exploit. As a result, any attribution method that relies on the computation of $\nabla_\x\ell(\x)$, e.g.,~\cite{AllConvNet,sundararajanIntegrtedGradients}, will necessarily be influenced by adversarial robustness.

Another important line of research in interpretability advocates using structured perturbations to explain neural networks' predictions~\cite{FongInterpretable,Fong_2019_ICCV} and to identify semantically relevant features of an input image. In particular, it has recently been proposed to use different types of image corruptions, such as deletion of objects, blurring, etc, to identify the regions of an image that can change, or preserve, the decision of a neural network. The techniques used in these works resemble those of adversarial attacks for non-additive perturbations~\cite{EykholtRobustPhysical,Wu2020Defending} as they also use similar type of corruptions to change the classifier's decision.

Counterfactual explanations~\cite{KarimiModel_Agnostic,SokolCounterfactural,Wachter_2017,SokolConversational} extend the previous ideas to general domains, and also propose to use semantically meaningful transformations to explain the decision of a classifier. In particular, we refer to the set of counterfactual explanations of a factual input $\x\in\R^D$ as the set 
\begin{equation}
    \mathcal{E}(\x)=\{\bm{x}'\in\mathcal{C} : f_\th(\x) \neq f_\th(\bm{x}')\},\label{eq:counterfactual}
\end{equation}
containing all input samples $\bm{x}'\in\mathcal{C}$ in a constraint set $\mathcal{C}$, for which the predictive model $f_\th$ makes a different prediction than in the factual input. Note the stark similarity between the set in \eqref{eq:counterfactual} and the definition of adversarial examples in \eqref{eq:adv_example}. Indeed, most methods for obtaining counterfactual explanations can be thought as adversarial attacks, and vice-versa. In the case of adversarial robustness, $\mathcal{C}$ usually represents some imperceptibility constraints, while in the field of counterfactual explanations it might represent some structured and meaningful perturbations, such as substituting one word in a sentence, or replacing one region of an image with some alternative content. Furthermore, unlike adversarial robustness, which has been mostly studied on continuous and differentiable spaces (mainly vision), the use of counterfactual explanations is mostly focused in domains with discrete and heterogeneous features and non-differentiable spaces. An example is the use of predictive models for supporting consequential decision making in contexts like pretrial bail and loan approval. Therefore, work at the intersection of both domains has an important opportunity for cross-fertilization~\cite{BalletImperceptible,yang_greedy_2020}.

Finally, it has recently been shown that, the same way that neural networks' decisions are vulnerable to adversarial attacks, most interpretability methods are also unstable under  perturbations of the input~\cite{Kindermans2019,ghorbani2019Fragile}. Indeed, it seems that one can construct adversarial perturbations which utterly change the interpretation of almost any attribution method. For this reason, some researchers are currently pursuing new  techniques, mostly inspired by adversarial defenses, to design more reliable and robust interpretations~\cite{Alvarez-Melis,ChenRobustAttr,ivankay2020far}.

\subsubsection{Robustness promotes interpretability}

At a high level, the robustness of a model to adversarial perturbations can be viewed as an invariance property satisfied by a classifier (see Section~\ref{sec:understanding_dl}). A network that can achieve a small loss within a set of admissible perturbations must therefore learn representations that are invariant to such perturbations. Hence, methods like adversarial training should be able to embed those invariances in a model. We suppose humans are invariant to these small perturbations, too. As a result, we can expect robust models to be more aligned with human vision, and consequently more interpretable than standard ones.

\begin{figure}[t]
    \centering
    \includegraphics[width=\columnwidth]{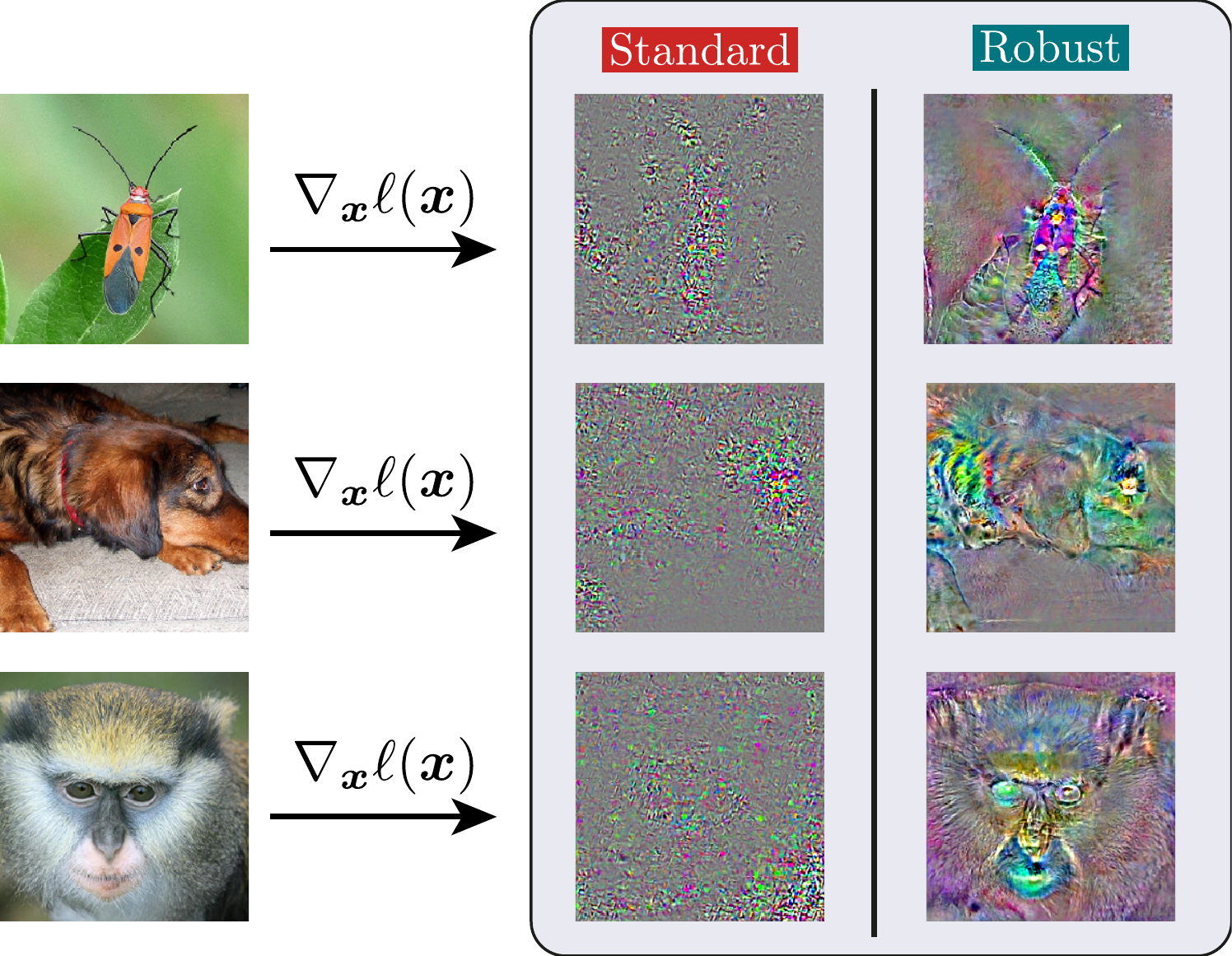}
    \caption{Comparison of the loss gradient with respect to the input, $\nabla_\x\ell(\x)$, of a ResNet-50 after standard training on ImageNet, and after adversarial training with $\ell_2$ adversarial examples crafted with PGD. Recall that $\nabla_\x\ell(\x)$ is the simplest way to highlight the input pixels that affect the classifier’s prediction the most. For the network trained on natural examples, the gradients appear very noisy. On the contrary, the gradients of the adversarially trained networks are more human-aligned, in the sense that they align better with perceptually relevant features. Image adapted from~\cite{tsiprasRobustnessMayBe2018} with permission from the authors.}
    \label{fig:better_saliency}
\end{figure}

\begin{figure}
    \centering
    \includegraphics[width=\columnwidth]{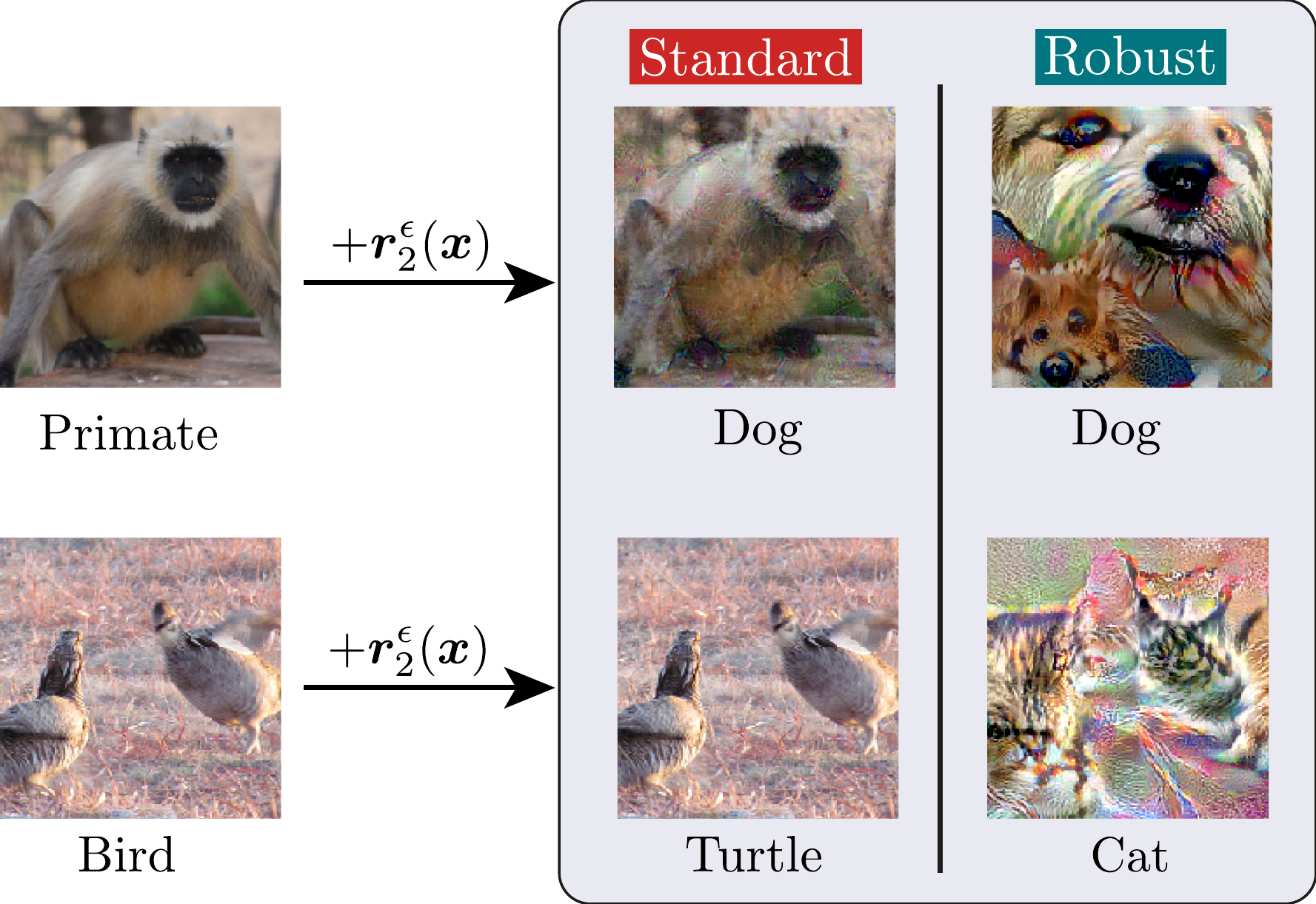}   \caption{$\epsilon$-constrained adversarial examples of a ResNet-50 trained standardly, and adversarially trained using $\ell_\infty$ and $\ell_2$ adversarial examples crafted with PGD, on ImageNet (the value of $\epsilon$ for such examples is much larger than the one used for training). These adversarial examples are constructed by iteratively following the negative loss gradient, $-\nabla_\x\ell(\x)$, of a desired/target class, i.e., ``dog'', while staying within an $\ell_2$-distance of $\epsilon$ from the original image. In contrast to standard models, for which adversarial examples appear as noisy variants of the original image, we can see that adversarial perturbations for robust models tend to produce salient characteristics of the fooling class. In fact, the corresponding adversarial examples can even be perceived as samples from that class. Image adapted from~\cite{tsiprasRobustnessMayBe2018} with permission from the authors.}
    \label{fig:high_epsilon_adv}
\end{figure}

Indeed, it has been shown in practice that, in contrast with standard models, the input gradients -- and saliency maps -- of adversarially trained networks align well with human perception~\cite{tsiprasRobustnessMayBe2018}. Figure~\ref{fig:better_saliency} shows an example of this, where we see how $\nabla_\x\ell(\x)$ highlights relevant features of an input image like the object shape, while in the standard model it looks more like random noise. Moreover, we also see how the adversarial examples computed on robust models\footnote{Note that even if one can certify that a model is robust to perturbations with a norm smaller than $\epsilon$ it is still possible to compute adversarial perturbations with a larger norm. These are of course perceptible.} do not look like meaningless noise, but instead tend to produce salient features of the class predicted by the model on the adversarial example. This effect is especially visible if the adversarial perturbations are allowed to have a very large norm (see Figure.~\ref{fig:high_epsilon_adv}).

Such level of perceptual alignment can be justified by the observation that the learned representations of robust models use primarily robust features of the training data (see Section~\ref{subsec:adversarial_robustness_and_generalization}). For this reason, some researchers argue that the robust models must learn salient features which are closer to our human cognition~\cite{ImageSynthesis,engstrom2019adversarialPriorLearnedRepresentations}, and therefore, that the local geometry of these models must reflect this alignment with a semantically meaningful $\nabla_\x\ell(\x)$. In fact, the apparent increase in interpretability seems to be a general property of robust models and not only of adversarially trained ones, as one can observe the same behaviour with models that use different methods to build robust models~\cite{rossImprovingAdversarialRobustness, kaur2019perceptuallyaligned}.

At this stage, it is important to highlight that the claims regarding human-alignment of robust models trained on vision data are mostly based on visualizations of their internal representations. In fact, to the best of our knowledge, no formal validation has been performed to assess the validity of this conjecture, yet. This verification should necessarily come from an extensive campaign of psychophysics experiments with human subjects, and would require a rigorous formalization of the notion of human alignment from a neuroscientific perspective\footnote{In the context of feature visualization, this type of studies have already been proposed to address the interpretability of different visualization  methods~\cite{borowski2020exemplary}.}. Preliminary work in this regard has already provided evidence in favour of some non-trivial connection between adversarial perturbations of standard models and human cognition~\cite{zhou_humans_2019}, but an in-depth study for robust models is still missing. Due to the importance of the human perspective in the use of robust models for downstream tasks, we see this formal verification as an important avenue of research for future studies.

\subsection{Transfer learning}\label{sec:transfer}
\label{subsec:transfer_learning}
Transfer learning~\cite{TorreyTransferLearning,TanTransferLearning} refers to the common practice in deep learning by which a neural network trained on one task (source) is adapted to boost performance on another related task (target). This practice is especially common in applications where data is scarce, or with low computational resources. In those cases, training a full deep neural network from scratch can be infeasible. Transfer learning aims to circumvent this problem and it proposes to use the weights of a trained network on the source task, to initialize the same architecture before training on the target task. Here, the main assumption -- which has been validated in practice~\cite{weiss_khoshgoftaar_wang_2016,huh2016makes,Kornblith_2019_CVPR} -- is that the learned representations of the source model can be translated, or transferred, to the target model via fine-tuning. In this way, one can leverage a large training set in the source task to identify good features, and simply exploit these learned features to boost the performance of the fine-tuned model on the target task. 

Recently, it has been discovered that the transfer learning performance of robust classifiers is better than that of standard models~\cite{salman2020adversarially, utrera2020adversariallytrained}. In this section, we review the main works explaining why this is the case, and practically leveraging this property in multiple applications.

\subsubsection{Robustness improves transfer learning performance}
\label{subsubsec:robustness_improves_transferability}
Among the great variety of transfer learning techniques, fine-tuning is probably one of the most widely adopted approaches, in which one uses a large model pre-trained on a source task (i.e.,~on ImageNet) to solve a target task. This is usually done either by fixing the parameters of some layers and fine-tuning only the rest using data of the target task; or by using the full pre-trained model as initialization, and fine-tuning the whole network on the new task.

Surprisingly, if instead of using a standard pre-trained model one uses its robust counterpart, the performance of fine-tuning transfer learning techniques improves significantly~\cite{salman2020adversarially,utrera2020adversariallytrained}. This happens despite the fact that robustly pre-trained models tend to have lower clean accuracy than standard ones. In general though, there exists a clear trade-off in terms of the requirements that a pre-trained model should satisfy in order to transfer well: higher robustness with lower accuracy on the original task improves transferability, but better accuracy with lower robustness does as well~\cite{salman2020adversarially,utrera2020adversariallytrained}. Nevertheless, it seems that this trade-off can be balanced if the robust network is adversarially trained using smaller perturbations~\cite{salman2020adversarially}. Consequently, smaller values of $\epsilon$ in \eqref{eq:epsilon_constraint} generally result in models that transfer better, a property that is in concordance with the findings presented later in Section ~\ref{subsubsec:ell_p_improves_common_corruptions}, where adversarial training with weaker attacks also promotes out-of-distribution generalization.

\begin{figure}
    \centering
    \includegraphics[width=\columnwidth]{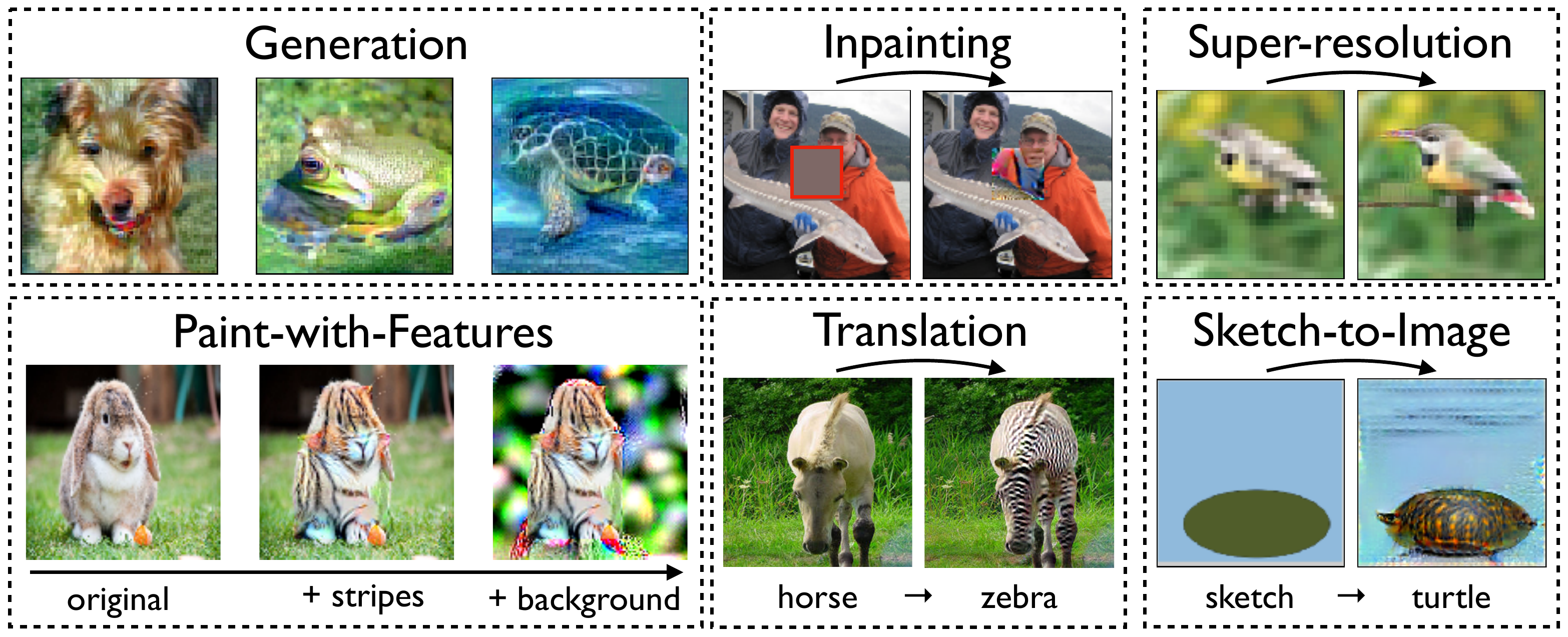}
    \caption{Examples of computer vision tasks, e.g., image synthesis or feature manipulation, solved using na{\"i}ve approaches with \emph{single robust} ResNet-50 classifier (adversarially trained on ImageNet using $\ell_2$ adversarial examples crafted with PGD). Because the learned representations of adversarially trained models encode more distinct characteristics of the underlying classes (cf.~Fig.~\ref{fig:high_epsilon_adv}), one can directly exploit them through gradient descent to precisely manipulate the input image features. By doing so, it is feasible to perform many complex, non-classification tasks, by using just a single robust classifier. Image taken from~\cite{ImageSynthesis} with permission from the authors.}
    \label{fig:image_synthesis}
\end{figure}

A possible reason for the enhanced transferability of robust models is the observation that they have learned feature representations that correlate better with semantically meaningful features of the input images~\cite{tsiprasRobustnessMayBe2018, ImageSynthesis, allen-zhu_feature_2020,engstrom2019adversarialPriorLearnedRepresentations}. In fact, the learned representations of adversarially trained classifiers are powerful enough to act as effective primitives for semantic image manipulations, unlike in standard models. This is despite the fact that they are solely trained to perform image classification~\cite{ImageSynthesis}. As a result, one can exploit the learned features of just a \emph{single} robust classifier to directly manipulate the input image features and perform complex tasks such as image generation and inpainting (see Figure~\ref{fig:image_synthesis}). These tasks have been mostly performed by techniques based on Generative Adversarial Networks (GAN)~\cite{GoodfellowGAN, YuInpaintingGAN,Ledig_2017_CVPR}, yet they work also well with a single robust classifier. The exact reason of why and how the feature representations of robust models enable such a broad exploitation is so far unknown.

\subsubsection{Robustness is transferable}
Looking deeper into the connection between robustness and transferability, it seems that one can actually reveal some further interesting properties. Not only adversarially trained models have the ability to improve the overall performance in transfer learning tasks, but it seems that they also lead to more robust models after fine-tuning. That is, \emph{robust features are also transferable}~\cite{HendrycksPreTraining, Shafahi2020Adversarially}. Even in cases where the performance (accuracy) of the fine-tuned models on the target task does not improve, their adversarial robustness improves significantly, sometimes reaching performance that is comparable to performing adversarial training from scratch on the target task~\cite{HendrycksPreTraining}. It is worth mentioning that the robustness transferability is not limited only to adversarial settings; the robustness of models trained to be robust to label corruptions and class imbalances also transfers well~\cite{HendrycksPreTraining}.

One possible explanation for this phenomenon could again be the fact that adversarially trained source models act as filters of non-robust features. Hence, removing the non-robust and noisy features, makes it easier for the fine-tuned models to find generalizable and robust features in the target dataset. 

Finally, beyond the idea that robust models act as filters (feature purification), the transferability of robustness can also be explained from a geometric perspective. Recall from Sections~\ref{sec:geometric_insights} and~\ref{sec:geometry_applications} that adversarial robustness changes the geometry of the learned classifier, an effect that is highly reflected in the input gradient $\nabla_\x\ell(\x)$. In that sense, it has also been shown that it is feasible to transfer the robustness of a source model by simply mimicking its geometric properties. Authors in~\cite{ChanWhatItThinks} proposed to enforce this by training a model on a target task with an additional term that matched the distribution of gradients of the robust source model. That is, during training, they penalized deviations between the distribution of target gradients $\nabla_\x\ell_{\text{T}}(\x)$, and the distribution of gradients $\nabla_\x\ell_{\text{S}}(\x)$ of the source network. They achieved this using some GAN-like objective function. The result of this strategy is a model with good performance on the target task but with enhanced robustness due to the fact that source and target gradient distributions are approximately equal. This observation highlights, once again, that the input gradients -- hence, the geometric properties of deep classifiers -- determine many characteristics of deep networks.

\subsection{Robustness to distribution shifts}
\label{sec:common_corruptions}

\begin{figure}
    \centering
    \includegraphics[width=\columnwidth]{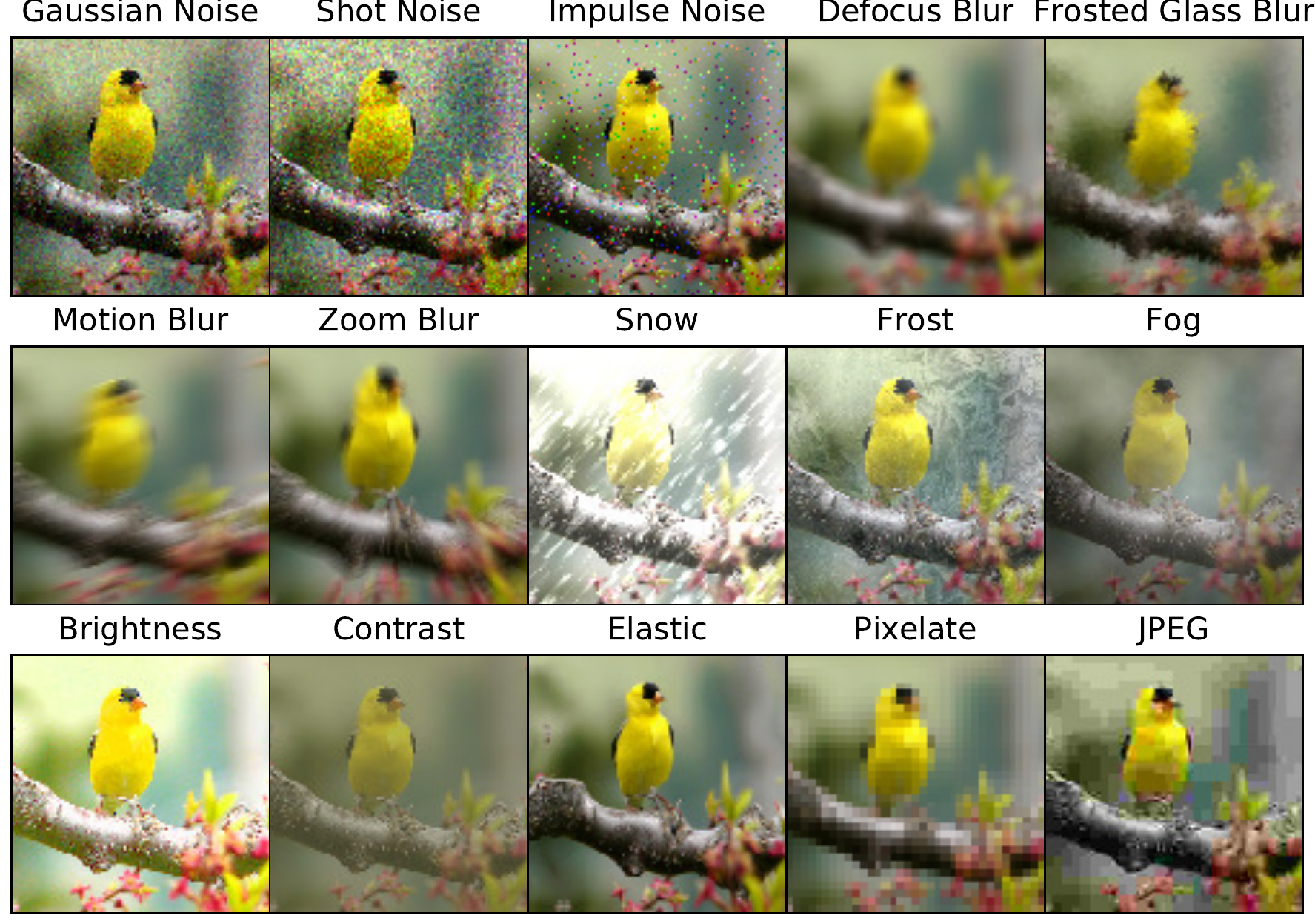}
    \caption{Examples of common corruptions applied to ImageNet, which can be used to benchmark the robustness of neural networks to different naturally occurring perturbations and distribution shifts. This corrupted version of ImageNet, called ImageNet-C~\cite{hendrycksBenchmarkingNeuralNetwork2019}, consists of $15$ types of algorithmically generated corruptions from noise, blur, weather, and digital categories. Each type of corruption has $5$ levels of severity, resulting eventually in $75$ distinct corruptions. Image taken from~\cite{hendrycksBenchmarkingNeuralNetwork2019} with permission from the authors.}
    \label{fig:examples_of_corruptions}
\end{figure}

Normally trained neural networks are brittle. Not only in the adversarial sense, but also their performance suffers when their test data is slightly different than the one used for training. In the machine learning community, the robustness to this type of general transformations is known as \emph{robustness to distribution shifts} or \emph{out-of-distribution generalization}. In contrast with the standard statistical learning framework (see Section~\ref{sec:dl_defs}), the notion of out-of-distribution generalization tries to circumvent the strict assumption of independently and identically distributed training and test data, and aims to train classifiers which generalize to test distributions that are slightly shifted, i.e., distant, from the training one. Naturally occurring shifts can be caused due to multiple reasons such as common corruptions~\cite{hendrycksBenchmarkingNeuralNetwork2019}, or dataset collection biases~\cite{RechtDoImagenet} (see Figure~\ref{fig:examples_of_corruptions}). It is widely believed that the vulnerability of neural networks to this type of shifts is due to their tendency to latch onto spurious correlations of the dataset, such as texture~\cite{geirhos2018TextureBias}, or background features~\cite{xiao2020noiseBG}, which might not generalize out of their narrow test distribution.

Unlike in the adversarial setting, where a proper formalization of the problem as in \eqref{eq:adv_example} and \eqref{eq:adv_accuracy} can be derived, the concept of out-of-distribution generalization encapsulates a more abstract notion of robustness. In this case, we are referring to robustness to any sort of change, imperceptible or not, random or adversarial, which retains the semantics of the input instances. As a result, although there exist some standard benchmarks to test the robustness of a network to a set of predefined class-preserving transformations~\cite{hendrycksBenchmarkingNeuralNetwork2019, hendrycks2020faces, hendrycks2019natural, geirhos2018TextureBias, RechtDoImagenet,recht2018cifar10}, evaluating the general robustness of a neural network to any naturally occurring perturbation is a hard problem. Adversarial robustness can act as a proxy for this task, allowing to compute lower bounds on the robustness of these systems, but also improve general robustness to a wide range of shifts when tuned properly. Furthermore, adversarial machine learning techniques can also be used to calibrate the confidence that a neural network assigns to samples outside of its training support, and hence allow for a better handling of out-of-distribution examples.

\subsubsection{Adversarial robustness as a lower bound}
\label{subsec:lower_bound}
When evaluated properly, the robustness of a neural network to adversarial perturbations can serve as a proxy to the robustness to some type of common corruptions. Indeed, if one can design an adversarial attack whose constraint set $\mathcal{C}$ and objective function $Q(\r)$ in \eqref{eq:adv_example} encapsulate a well-defined set of perturbations, the adversarial robustness of a neural network to this attack will act as a lower bound to the robustness to all perturbations in $\mathcal{C}$, i.e., a measurement of the worst-case performance of the system.

Take the robustness of a neural network to random noise as an example, and let $\r^{\bm{v}}_2(\x)$ denote the additive perturbation required to move a sample $\x$ to the decision boundary of a neural network in the direction given by the unit vector $\bm{v}\in\mathbb{S}^{D-1}$, i.e.,
\begin{equation}
    \r_2^{\bm{v}}(\x)=\underset{\r\in\{\alpha \bm{v}:\alpha\in\R\}}{\operatorname{arg\;min}} \|\r\|_2\text{ s.t. }f_\th(\x+\r)\neq f_\th(\x).
\end{equation}
It can been shown theoretically that both for linear~\cite{Fawzi_Analysis_MachineLearning}, and non-linear classifiers with low local curvature~\cite{fawziRobustnessClassifiersAdversarial2016}, such as neural networks, the norm of the random noise $\|\r^{\bm{v}}_2(\x)\|_2$ required to fool the classifier is
\begin{equation}
    \|\r^{\bm{v}}_2(\x)\|_2=\Theta\left(\sqrt{D}\|\r_2^\star(\x)\|_2\right)
\end{equation}
with high probability. This means that improving the adversarial robustness to $\r_2^\star(\x)$ perturbations will provably increase its robustness to random noise\footnote{Note also that in the high-dimensional setting, as it is common in deep learning, the robustness of a neural network to random noise can be orders of magnitude higher than that to $\r_2^\star(\x)$ perturbations. As seen in Section~\ref{sec:geometric_insights}, this is a result of the low mean curvature of deep classifiers.}\textsuperscript{,}\footnote{Recently, some authors have argued that the sense of causation is in fact the opposite~\cite{gilmer19aNaturalConsequencTestError}. Namely, that the vulnerability of neural networks to random noise in a high-dimensional setting is responsible for the extreme vulnerability of neural networks to adversarial perturbations. In this view, adversarial examples are a natural consequence of the finite robustness to random noise, and hence it is the latter phenomenon the one that really needs to be addressed.}. 

Nevertheless, each specific formulation of \eqref{eq:adv_example} based on a choice of $Q(\r)$ and $\mathcal{C}$ can only guarantee robustness to one narrow type of perturbations, e.g., for $\r^\star_2(\x)$ we can lower bound the robustness to Gaussian isotropic noise. Evaluating the robustness of deep classifiers to more types of corruptions requires, therefore, to test using other types of adversarial perturbations. However, for most common corruptions, we still do not have a proper mathematical characterization of the set of plausible perturbations. In these cases, the misalignment between the metrics used to compute adversarial examples and the distribution of naturally occurring shifts renders the evaluation of the worst-case robustness against these data transformations impossible. In fact, this metric disagreement might also be responsible for the low performance of $\ell_p$-robust networks in certain natural distribution shift benchmarks. In these benchmarks, improving $\ell_p$ robustness does not improve accuracy~\cite{Taori2020}. Improving the performance on this task will, therefore, require a different design of the adversarial constraints and the objective metric. As we will see in Section~\ref{subsec:beyond_lp}, this is an ongoing research effort in the adversarial robustness community.

\subsubsection{Adversarial $\ell_p$ robustness improves robustness to common corruptions} \label{subsubsec:ell_p_improves_common_corruptions}
In practice, however, adversarially trained models against $\ell_p$ perturbations seem to be more robust to a wide range of distribution shifts than their standardly trained counterparts. They show a smaller gap between their clean performance and their performance under common corruptions~\cite{hendrycksBenchmarkingNeuralNetwork2019}, and seem to be less biased towards spurious attributes such as image texture~\cite{geirhos2018TextureBias}. Furthermore, adversarially trained models are more robust to perturbations that alter the high-frequency components of image data~\cite{yinFourierPerspectiveModel2019,sharmaEffectivenessLowFrequency,LiuQFool,tsuzukuStructuralSensitivityDeep2019,RahmatiGeoDA,ortiz-jimenezHoldMeTight2020}. This is conjectured to be a result of the better alignment between the features exploited by robust models and human perception~\cite{Xie_Improve}.

Yet, although they perform better than standard models, the models that are adversarially trained using strong security budgets, e.g., large-$\epsilon$ constraints in \eqref{eq:epsilon_constraint}, still perform poorly on the common corruption benchmarks. Nevertheless, recently some researchers have argued that some adversarial robustness techniques can still be used to improve out-of-distribution generalization~\cite{hendrycksBenchmarkingNeuralNetwork2019,geirhos2018TextureBias,Xie_Improve}.

In particular, adversarial training can be seen as a form of data augmentation~\cite{bochkovskiy2020yolov4,TangOnlineAugment}, in which a neural network learns to distinguish between the useful generalizing features and the spurious correlations of the training data. In this regard, properly-tuned adversarial training schemes can be exploited to improve the out-of-distribution generalization capacity of the network, by forcing it to learn better invariances of the input data. The key to the success of these methods resides in the choice of adversarial perturbations crafted during adversarial training. In particular, because their aim is not improving security to a specific attack, but rather the general robustness of the models, most of these techniques exploit weaker attacks~\cite{Xie_Improve,TangOnlineAugment}. Their empirical results suggest that this is enough to drive the networks to learn human-aligned representations similar to those discussed in Section~\ref{subsec:robust_features}, improving the general robustness of these systems to some naturally occurring shifts~\cite{hendrycksBenchmarkingNeuralNetwork2019, hendrycks2020faces, hendrycks2019natural, geirhos2018TextureBias}, albeit at the cost of worse adversarial robustness.

The idea of exploiting adversarial training to improve representations is not new in natural language processing (NLP). Indeed, in this field it was early recognized that adversarial training could help avoid common pitfalls of language models by preventing neural networks from learning spurious correlations in the data~\cite{miyato2016adversarial,Miyato2019,cheng-etal-2019-robust,Zhu2020FreeLB}. In most of these works, adversarial training is not performed on the input space -- which in NLP would mean to flip a character or change a word -- but on the embedding space. This way, the networks are forced to learn important invariances in the semantically meaningful embedding space and hence to avoid overfitting. 

\subsubsection{Beyond the $\ell_p$ metric}
\label{subsec:beyond_lp}

The clear misalignment between the $\ell_p$ metrics and naturally occurring perturbations, however, makes researchers doubt that larger performance gains can be achieved in the more general robustness problem by only improving the performance on $\ell_p$ adversarial tasks~\cite{Taori2020}. For this reason, a plethora of works has recently started to arise in which new types of adversarial attacks beyond the $\ell_p$ metric are being introduced. Most of these attacks modify either the constraint set $\mathcal{C}$ or the objective function $Q(\r)$ in \eqref{eq:adv_example} such that the resulting adversarial perturbations resemble naturally occurring data transformations that might go beyond imperceptibility or additivity. Most of these perturbations have been studied in the context of image data and address transformations such as rotations, translations and shears~\cite{fawziManitestAreClassifiers2015,kanbakGeometricRobustnessDeep2018,EngstromExploring, xiao2018spatially,athalye_synthesizing_2018, alaifari2019adef, Alcorn2019Pose}, changes in color~\cite{Laidlaw_Functional,Shamsabadi_ColorFool}, or structurally meaningful perturbations as measured by the Wasserstein metric~\cite{WongWasserstein,WuWY20} or generative models~\cite{fawziAdversarialVulnerabilityAny2018,laidlaw2020perceptual,wong2020learning}.

The main problem with the above approaches is that most formulations of adversarial perturbations that do not exploit the $\ell_p$ metric, or that are not additive, are generally hard to solve. This is mostly due to the inefficacy of standard first-order optimization methods in these cases, generally with high computational complexity requirements. Coming up with metrics that can better capture common corruptions for different types of data, and for which adversarial attacks can be efficiently computed, is, still to this date, an important challenge for the research community.

\subsubsection{Adversarial robustness for confidence calibration}
Orthogonal to the previous approaches, another line of research tackles the overconfidence of neural networks outside of their training distribution to improve their behaviour under distribution shifts. Recall from Section~\ref{sec:dl_defs} that $[p_\th(\x)]_k\in[0,1]$ represents the probability estimated by a neural network that the input $\x$ should be assigned the label $k$. Nevertheless, this value is poorly calibrated in most neural networks, as it is generally very high regardless of the data sample. In particular, we say that a neural network is overconfident when for random samples outside the support of its data distribution\footnote{For image data, this could be a sample with a random pattern of pixel intensities.}, $\bm{x}'\nsim\mathcal{D}_{\x}$, the probability assigned to the network's prediction is very high, $\max_k [p_\th(\bm{x}')]_k\approx 1$. Properly calibrating the confidence of a neural network outside the support of their input data is important because it allows to detect inputs for which the network might be uncertain. This can help flag out-of-distribution samples and hence avoid potentially harmful decisions of AI systems.

An efficient way to reduce the overconfidence of neural networks is actually to use adversarial training to maximize the uncertainty of a deep classifier outside of the support of its training distribution. For example, adversarial confidence enhanced training (ACET)~\cite{Hein_ReLUConfidence} is a method that can alleviate this problem, as, during training, it approximately solves
\begin{equation}
    \min_\th \mathbb{E}_{(\x,y)\sim\mathcal{D}}\left[\L(\x,y;\th)\right]+\lambda \mathbb{E}_{\bm{x}'\nsim\mathcal{D}_{\x}}\left[\max_{k\in\{1,\dots,C\}}[p_\th(\x')]_k\right].
\end{equation}
Because approximating the expectation over $\bm{x}'\nsim\mathcal{D}_{\x}$ using finite samples is statistically inefficient, ACET uses an adversarial approach and at every step of SGD, for every sample $\bm{x}'\nsim\mathcal{D}_{\x}$, it adds
\begin{equation}
    \max_{\bm{u}\in\R^D} \max_{k\in\{1,\dots,C\}}\;[p_\th(\bm{u})]_k\text{ subject to }\|\bm{x}'-\bm{u}\|_p\leq\eta,
\end{equation}
to the objective function. This term forces the network to minimize the confidence $\max_{k\in\{1,\dots,C\}}[p_\th(\bm{u})]_k$ on the sample with the largest current confidence in an $\eta$-ball around non-typical examples. This trick improves the statistical efficiency of ACET, because it guarantees that the overconfidence of the neural network is always minimized in the worst-case sense, instead of on-average. Other works~\cite{Zou_2019_ICCV,StutzCCAT} have also proposed extensions of adversarial training to reduce overconfidence and improve robustness to unseen attacks, such as assigning soft labels with low confidence to the perturbed samples, and even directly rejecting those adversaries that already have low confidence in the original model. This way, the network does not overfit to the current attack and it is able to learn a function that can reject perturbations outside of those in its training set.

\subsection{Other applications of adversarial robustness}
The applications of adversarial robustness reviewed above are just fraction of the spectrum of topics that have adopted adversarial machine learning techniques to tackle some of their challenges. In this sense, many problems in deep learning can be framed in terms of geometric properties of neural networks, and as such, they can benefit from the wide range of adversarial robustness techniques that can modify the network's geometry. For example, in the context of anomaly detection, adversarial perturbations have been used to generate synthetic anomalous samples which do not belong to the data manifold~\cite{liang2018enhancing, GoyalDROC}. In data privacy, adversarial techniques have also been exploited, for example, to identify sensitive features of a sample which can be vulnerable to attribute inference attacks~\cite{attriguard}. However, it has also recently been argued that adversarial robust models are more vulnerable to membership inference attacks~\cite{songMembership}. Finally, in the context of fairness, adversarial machine learning has recently found multiple applications~\cite{Yurochkin2020Training,edwards2015censoring,GargCounterfactual,ZhangMitigating,madras18a}, from the use of adversarial training to favour invariance towards racially-, or genderly-, biased features~\cite{Yurochkin2020Training}; to the formulation of different notions of fairness, such as the right to an equally robust machine learning prediction~\cite{fairness_through_robustness}. 

Overall, it is clear that adversarial robustness has many applications beyond securing systems to adversarial attacks. In particular, the ability of adversarial defenses to induce invariance to certain transformations of the data, and hence, learn better representations of its features, is making adversarial robustness a fundamental piece of the deep learning toolbox.

\section{Future research and open questions}
\label{sec:future}

Since the discovery of adversarial examples in deep learning, adversarial robustness has become a very active research field. It is transforming our theoretical understanding of deep neural networks (Section~\ref{sec:understanding_dl}), while it has also become a fundamental tool for many applications of deep learning beyond security (Section~\ref{sec:applications_robustness}). Nevertheless, despite the great achievements in building more robust, effective, and explainable architectures in the recent years, there are still some exciting open questions and challenges in the field.

For example, despite years of research about $\ell_p$ robustness, the main current challenge is still coming up with methods that achieve substantial levels of robustness at a lower cost in terms of standard generalization performance. Whether it is possible to obtain robust models that perform on par with standard ones in terms of accuracy, even in the strongest attack setting, is still an open problem. 
In particular, recent results suggest that trading off a bit of robustness~\cite{ZhangTRADES, lamb_interpolated_2019}, by letting networks be slightly more susceptible to adversarial attacks, or by exploiting unlabeled data~\cite{CarmonUnlabeledAT,RaghunathanUnderstanding,AlayracAreLabelsRequired,Taori2020} during training to increase the statistical efficiency of the learnt invariances, can yield substantial gains in the networks' performance.

In a more theoretical note, the theory of adversarial robustness and its connection to statistical learning theory is largely unexplored. Only recently, some authors have realized that the main mathematical tool in statistical learning theory, i.e., uniform convergence, might not be the right notion to study the generalization performance of deep neural networks~\cite{nagarajanUniformConvergenceMay2019}. The reason for this is that uniform convergence bounds can become vacuous for classifiers for which in-distribution adversarial examples exist. Factoring in the effect of adversarial examples in the development of new generalization bounds for deep learning is, therefore, an important line of research.

Then, in relation to constructive methods, the important question of how and why do standardly trained neural networks choose, and prefer, non-robust features of a dataset remains widely open. Recent works have shown that neural networks possess a specific signature in the form of a set of vectors that encodes their own preference to discriminate the data based on some particular features. These are coined as the neural anisotropy directions (NADs) of an architecture~\cite{ortizjimenez2020neural}. An important challenge is, however, to understand how these signatures are encoded in the architecture and then to devise ways to steer them towards robust features. In this sense, some researchers are trying to come up with architectural improvements that can boost robustness, such as smoothing the loss landscape with new activation functions~\cite{xie2020smoothAT}, or introducing robustness objectives in large-scale neural architecture search loops~\cite{GuoNASRobustness,dong_adversarially_2020}. Besides, recent works have proposed to leverage our rich knowledge of the human visual system~\cite{serre_deep_2019} to design architectures that can directly filter out non-robust features~\cite{DapelloMonkey,zhou_humans_2019,reddy2020biologically}. However, most of these methods have mainly been tested in the small-$\epsilon$ regime and, hence, their performance has not yet been compared to the one of standard adversarial defenses, e.g., adversarial training, which are evaluated using higher values of $\epsilon$.

Furthermore, it seems that adversarial training is a powerful technique that allows neural networks to learn features which are better aligned with human perception. However, obtaining models that are robust to a diverse set of naturally occurring distribution shifts stays an open problem. Leveraging our knowledge in adversarial robustness of deep learning models seems to offer a promising way forward. Yet, our current techniques are still falling short to solve this very general problem. Some studies even suggest that our recent improvements in adversarial robustness have not yielded any improvements over the capacity of our models to generalize to small distribution shifts, which do not alter human baselines~\cite{RechtDoImagenet,EvaluatingImageNet, Taori2020}. They argue that, as a result, our current systems' accuracy in the real world is much lower than the one reported in their validation sets. For this reason, more research is necessary to address the fragility of deep learning to small distribution shifts. Nonetheless, it is clear that the design of new adversarial attacks that depart from the typical $\ell_p$ settings, is a promising avenue for future research in evaluating, and improving, the robustness and out-of-distribution generalisation in more general settings.

Also, it is important to emphasize that adversarial training is a computationally expensive procedure. For this reason, despite its ability to learn more robust representations, it is not widely adopted outside of the computer vision applications. Specifically, the NLP community is currently experiencing an ongoing trend towards exploiting large models trained on massive datasets~\cite{gpt3}. For these applications, it is computationally challenging to perform adversarial training, which explains why this form of training is not widely adopted in NLP. Coming up with faster and more efficient adversarial training techniques that are tailored for these larger-scale applications is, therefore, a major challenge for future work.

Finally, it is important to acknowledge how, despite its rapid advancements in recent years, the focus of the adversarial robustness community on the attack-defense cycle has prevented researchers from realizing that there is ample room to apply adversarial robustness in fields beyond security. In this sense, many applications of machine learning benefit from learning rich invariances from data. Nevertheless, if we only judge the performance of robust models based on their vulnerability to certain adversarial attacks, e.g., $\r_\infty^\epsilon(\x)$ with a specific value of $\epsilon$, we might overlook some other aspects of robustness beyond merely security concerns. As we saw in Section~\ref{sec:applications_robustness}, a ``good'' model is not always the one that scores higher in the standard robust benchmarks. Hence, an over-reliance on these metrics for measuring the advancements of the field might be counterproductive. In general, it is important that the research on adversarial robustness should be steered towards defining novel set of benchmarks that also test the performance of adversarially robust models in other applications such as transfer learning, interpretable machine learning, and image generation.

\section{Concluding remarks}
\label{sec:conclusions}
In this article, we have provided an introduction to the field of adversarial robustness focused on its intersection with other deep learning fields. In this sense, in stark contrast with the mainstream perspective, we have steered our focus away from a security angle, and reviewed the main uses of adversarial robustness as a tool to understand and improve deep learning. 

Beginning with an overview of the main notions in adversarial robustness, we have seen how adversarial perturbations are intimately connected to the geometry of the decision boundary of a neural network, and we have explained why and how the local geometry of a deep classifier determines many of its properties. The study of the adversarial examples of a neural network, therefore, sheds light on the fundamental principles of deep learning. We have outlined the main works, in this regard, and we have shown how the adversarial examples of a network are a reflection of non-robust features it uses to discriminate the data. In this sense, we have seen how adversarial training acts as a filtering procedure that drives a neural network towards more robust solutions. Furthermore, we have seen how tracking the evolution of the decision boundary of a neural network during training can reveal important information about its learning dynamics.

In addition to improving our understanding of deep learning models, adversarial robustness has found many applications in deep learning. In this article, we have attempted to summarize most of these, and provided an integrated perspective on their guiding principles. In particular, we have highlighted how learning richer invariances of the input data using adversarial robustness techniques leads to better feature representation. Some of the best performing models in interpretability, transfer learning, or out-of-distribution generalization, exploit these to excel in a plethora of tasks. 

Although the field of adversarial robustness has rapidly evolved in recent years, its way ahead is far from over. The field is, currently, undergoing an important transformation as it moves away from a negative view of adversarial examples, to an optimistic perspective over the uses of robust models beyond security. In this sense, recent advances in our understanding of deep learning have expanded new applications of robust neural networks, unimaginable some time ago. However, it is clear that the field itself has only
scratched the surface of what is possible, and it will surely face many new problems, discoveries, and insights in the future.

\appendices

\ifCLASSOPTIONcaptionsoff
  \newpage
\fi

\bibliographystyle{IEEEtran}
\bibliography{IEEEabrv,main.bib}

\end{document}